\documentclass{article}

\usepackage[final, nonatbib]{neurips_2022}

\usepackage[utf8]{inputenc} 
\usepackage[T1]{fontenc}    
\usepackage{hyperref}       
\usepackage{url}            
\usepackage{booktabs}       
\usepackage{amsfonts}       
\usepackage{nicefrac}       
\usepackage{microtype}      
\usepackage{xcolor}         

\usepackage{enumitem}
\usepackage[pdftex]{graphicx}
\usepackage{grffile}
\usepackage{amsmath}
\usepackage{amssymb}
\usepackage{amsfonts}
\usepackage[bb=dsserif]{mathalpha}
\usepackage{algorithm}
\usepackage{algorithmic}
\usepackage{colortbl}
\usepackage{multirow}
\usepackage{multicol}
\usepackage{wrapfig}
\usepackage{caption}

\newcommand\eref{Eq.~\ref}
\newcommand\fref{Figure~\ref}
\newcommand\tref{Table~\ref}
\newcommand\sref{Section~\ref}
\newcommand\ha{ \rowcolor{orange!20}}

\def\loss{{\mathcal{L}}}
\newcommand\attn{\text{Att}}
\DeclareMathOperator*{\argmax}{arg\,max}
\DeclareMathOperator*{\argmin}{arg\,min}

\title{A Fast Post-Training Pruning Framework for Transformers}

\author{%
  Woosuk Kwon\thanks{Equal contribution.} \\
  UC Berkeley \\
  \tt{woosuk.kwon@berkeley.edu} \\
  \And
  Sehoon Kim$^\ast$ \\
  UC Berkeley \\
  \tt{sehoonkim@berkeley.edu} \\
  \And 
  Michael W. Mahoney \\
  UC Berkeley, ICSI, \& LBNL \\
  \tt{mahoneymw@berkeley.edu} \\
  \And
  Joseph Hassoun \\
  Samsung Semiconductor, Inc. \\
  \tt{j.hassoun@samsung.com} \\
  \And
  Kurt Keutzer \\
  UC Berkeley \\
  \tt{keutzer@berkeley.edu} \\
  \And
  Amir Gholami \\
  UC Berkeley \\
  \tt{amirgh@berkeley.edu}
}

\begin{document}

\maketitle

\setcounter{footnote}{0}
\begin{abstract}
Pruning is an effective way to reduce the huge inference cost of Transformer models.
However, prior work on pruning Transformers requires retraining the models.
This can add high training cost and high complexity to model deployment, making it difficult to use in many practical situations.
To address this, we propose a fast post-training pruning framework for Transformers that does not require any retraining.
Given a resource constraint and a sample dataset, our framework automatically prunes the Transformer model using structured sparsity methods.
To retain high accuracy without retraining, we introduce three novel techniques: 
(i) a lightweight mask search algorithm that finds which heads and filters to prune based on the Fisher information;
(ii) mask rearrangement that complements the search algorithm; and
(iii) mask tuning that reconstructs the output activations for each layer.
We apply our method to BERT$_\textsc{BASE}$ and DistilBERT, and we evaluate its effectiveness on GLUE and SQuAD benchmarks.
Our framework achieves up to 2.0$\times$ reduction in FLOPs and 1.56$\times$ speedup in inference latency, while maintaining $<$ 1\% loss in accuracy.
Importantly, our framework prunes Transformers in less than 3 minutes on a single GPU, which is over two orders of magnitude faster than existing pruning approaches that retrain the models.\footnote{Our code is publicly available at \tt\href{https://github.com/WoosukKwon/retraining-free-pruning}{https://github.com/WoosukKwon/retraining-free-pruning}}
\end{abstract}

\section{Introduction}
\label{sec:intro}

In recent years, Transformer~\cite{vaswani2017attention} has become a \textit{de facto} standard model architecture in Natural Language Processing~\cite{brown2020language,devlin2018bert,liu2019roberta},
and it is becoming common in many domains including Computer Vision~\cite{dosovitskiy2020image,liu2021swin,touvron2021training} and Speech Recognition~\cite{baevski2020wav2vec,chen2021wavlm,hsu2021hubert}.
However, efficient deployment of Transformer architectures has been challenging due to their large model size and high inference latency.
As a promising way to tackle this challenge, structured pruning of Transformers has been widely studied.

While prior work on pruning Transformers substantially reduces the inference time, it is often difficult to use in practice for several reasons.
First, previous approaches require retraining the pruned model and/or jointly learning the pruning configurations during training.
This increases the training time by up to 10$\times$~\cite{lagunas2021block, xia2022structured}, adding significant computational overhead.
Second, previous methods add many moving parts to the model deployment process.
That is, the pruning pipelines are often complex and require additional hyperparameter tuning.
Such techniques demand significant engineering efforts for implementation and debugging, which impedes their adoption in production pipelines.
Third, these previous methods do not directly adapt to the users' constraints.
They either rely on vague regularization hyperparameters to control the model sparsity or use fixed model architectures selected independently of the user settings.
This can result in sub-optimally pruned models that are not tailored to the users' constraints and hardware.

\begin{figure}[t]
\centering
    \includegraphics[width=0.8\textwidth]{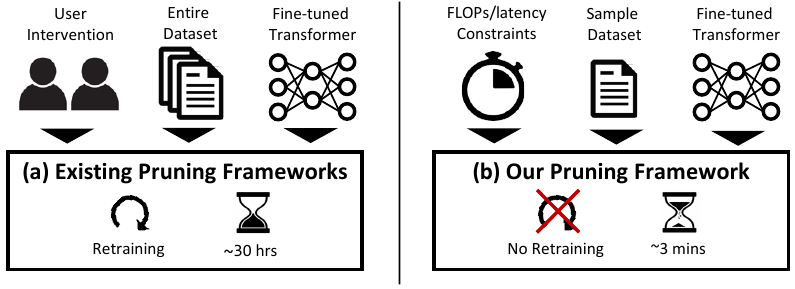}
\caption{
(a) Prior pruning frameworks require additional training on the entire training set and involve user intervention for hyperparameter tuning.
This complicates the pruning process and requires a large amount of time (e.g., $\sim$30 hours).
(b) Our pruning framework does not require retraining. 
It outputs pruned Transformer models satisfying the FLOPs/latency constraints within considerably less time (e.g., $\sim$3 minutes), without user intervention.
}
\vspace{-5mm}
\label{fig:overview}
\end{figure}

To address the above limitations, we propose a fast \textit{post-training pruning} framework for Transformers that does not require any retraining of the models.
As illustrated in \fref{fig:overview}, our framework takes as input a Transformer model, a sample dataset, and a FLOPs/latency constraint.
It then outputs a pruned Transformer model that can be deployed immediately.
By avoiding expensive retraining, the end-to-end pruning pipeline can be extremely fast and simplified, which typically takes a few minutes without any user interventions that complicate the whole process.

Indeed, post-training compression has been widely studied for quantization, and it gained considerable attention in both academia and industry~\cite{banner2018post,hubara2020improving,zhao2019improving}. 
Although quantization-aware training methods
achieve higher compression rates in general, post-training quantization (PTQ) has often been more preferred in practice due to its retraining-free advantages.
Importantly, PTQ allows quantization to happen seamlessly at the model deployment time using the tools such as TensorRT~\cite{tensorrt}, TFLite~\cite{tflite}, and OpenVINO~\cite{openvino}.
Similar to the PTQ methods, our framework provides an out-of-the-box tool that enables pruning of Transformers without engineering~efforts.

Our contributions can be summarized as follow:
\vspace{-2mm}
\begin{itemize}[leftmargin=*]
    \item We propose a novel post-training pruning framework for Transformers that does not require model retraining.
    To retain accuracy without retraining, our framework consists of three stages:
    (i) the \textit{mask search} process guided by the Fisher information matrix to select which heads/filters to prune (\sref{subsec:search});
    (ii) the \textit{mask rearrangement} process that reselects the heads/filters to prune by capturing intra-layer interactions (\sref{subsec:rearrange});
    and (iii) the \textit{mask tuning }process that adjusts the mask variables to ensure that the output signal is recovered for each layer (\sref{subsec:tuning}).
    
    \item We extensively test our framework by applying it to BERT$_\textsc{base}$ and DistilBERT on GLUE and SQuAD tasks (\sref{subsection:performance}).
    Within 1$\%$ of accuracy drop, our framework reduces 30--50$\%$ of the original FLOPs (\fref{fig:flops_result}), resulting in up to 1.56$\times$ speedup on an NVIDIA V100 GPU (\tref{tab:latency_result}).
    \item We show that our method achieves comparable or even better FLOPs-accuracy trade-off than prior structured pruning methods \textit{without} retraining (\sref{subsection:comparison}, \fref{fig:comparison}).
    Our end-to-end pruning pipeline finishes in only 39 and 135 seconds on average for GLUE and SQuAD (\sref{subsection:analysis}, \tref{tab:time_breakdown}), which is over 100$\times$ faster than the retraining-based methods.
\end{itemize}

\vspace{-2.5mm}
\section{Related Work}
\label{sec:related}
\vspace{-1mm}

\textbf{Efficient Transformers.} 
In order to improve the inference speed and reduce the memory footprint of Transformers, multiple different approaches have been proposed.
These can be broadly categorized as follows:
(i) efficient architecture design~\cite{iandola2020squeezebert,kitaev2019reformer, lan2019albert,sun2020mobilebert, wang2020linformer,wu2020lite};
(ii) hardware-software co-design~\cite{ham20203, ham2021elsa, tambe2021edgebert, wang2021spatten};
(iii) knowledge distillation~\cite{jiao2019tinybert, sanh2019distilbert,sun2019patient, wang2020minilm};
(iv) quantization~\cite{kim2021bert, shen2020q, zadeh2020gobo, zafrir2019q8bert};
(v) neural architecture search~\cite{chen2020adabert, so2019evolved, so2021primer,  wang2020hat, xu2021bert, yin2021autotinybert};
and (vi) pruning.
In this paper, we focus on pruning and briefly discuss the related works.

\textbf{Transformers Pruning.}
Pruning has been a promising way to remove unimportant weights in neural networks.
Pruning can be largely categorized into unstructured and structured pruning. 
For unstructured pruning, magnitude-based~\cite{gale2019state}, first-order~\cite{sanh2020movement}, and second-order~\cite{kurtic2022optimal} pruning methods, and the lottery ticket hypothesis~\cite{chen2020lottery, chen2020earlybert, frankle2018lottery, prasanna2020bert} have been explored for Transformers.
While these methods can substantially compress the model size, commodity hardware such as GPUs can hardly take advantage of the unstructured sparse patterns for model inference speedup.

For this reason, a number of structured pruning methods have been introduced to remove coarse-grained sets of parameters in Transformers.
For example, to prune structured sets of parameters in weight matrices, low-rank factorization~\cite{wang2019structured}, block-wise sparsity~\cite{li2020efficient}, and tile-wise sparsity~\cite{guo2020accelerating} were studied.
Furthermore, as more coarse-grained methods, attention head pruning~\cite{michel2019sixteen, voita2019analyzing} and layer dropping~\cite{fan2019reducing, sajjad2020effect} have been popularly used.
Taking a step further, recent approaches~\cite{chen2021chasing, khetan2020schubert,  lagunas2021block, lin2020pruning, liu2021rosita, xia2022structured, yao2021mlpruning} have explored jointly pruning Transformers with different pruning granularity and principles, maximizing the model efficiency in every dimension.
Orthogonally, another thread of work~\cite{fan2019reducing, hou2020dynabert, liu2021ebert, xin2020deebert, zhou2020bert} has shown that Transformers can be dynamically pruned at inference time.

Unfortunately, while the structured pruning methods can achieve high compression rates and speedups, they are often difficult to use in practice.
One reason for this is the high computational cost of additional training during or after pruning, which can be up to 10$\times$~\cite{lagunas2021block, xia2022structured} compared to that of the original model training.
Another reason is the high complexity of the pruning pipelines~\cite{hou2020dynabert, lagunas2021block, liu2021rosita, yao2021mlpruning}, where each pruning stage often requires rewriting the training code and introduces additional hyperparameters to tune.

\textbf{Post-training Model Compression.}
Post-training compression methods have been widely studied in quantization.
These methods, categorized as post-training quantization (PTQ), perform quantization without any retraining, thereby avoiding the additional training cost and user intervention.
Multiple PTQ techniques have been proposed to effectively mitigate the accuracy degradation without retraining~\cite{banner2018post, hubara2020improving, nagel2020up, zhao2019improving}.
 
Although not as much as for quantization, post-training schemes have also been explored for unstructured~\cite{frantar2022spdy, lazarevich2021post, mussay2019data} and structured pruning of CNNs.
For structured pruning, \cite{kim2020neuron, srinivas2015data, yvinec2021red} proposed ways to group and merge similar neurons in a CNN.
However, we find it difficult to extend those techniques to pruning Transformers because they require the model to have a repeating structure of a linear layer and an element-wise nonlinearity, which is not the case for the multi-head attention layers of Transformers.
Even for the feed-forward network layers of Transformers,
\cite{kim2020neuron, srinivas2015data} can hardly be used because they rely on a certain characteristic of ReLU while many Transformers~\cite{brown2020language, devlin2018bert, yang2019xlnet} use GELU~\cite{hendrycks2016gaussian} instead of ReLU.

Motivated by the fact that the existing post-training CNN pruning techniques cannot be applied to Transformers, in this paper we propose a novel post-training pruning method with a focus on Transformers.
However, we would like to note that the underlying principles in our approach are general enough to be extended to pruning other types of model architectures such as CNNs.

\vspace{-2mm}
\section{Overview}
\label{sec:overview}

\vspace{-2mm}
\subsection{Background}
\label{subsec:background}
\vspace{-1mm}

\textbf{Transformer Architecture.}
In this paper, we focus on the pruning of encoder-based Transformer~\cite{vaswani2017attention} models, especially the BERT~\cite{devlin2018bert} architecture family.
BERT is a stack of homogeneous Transformer encoder blocks, each of which consists of a multi-head attention (MHA) layer followed by a point-wise Feed-Forward Network (FFN) layer.
Specifically, an MHA layer consists of $H$ independently parameterized attention heads:
\vspace{-2mm}
\begin{align*}
\text{MHA}(\text{x}) &= \sum_{i=1}^{H} \attn_{i}(\text{x}), \quad
\text{x}_\textsc{{mha}} = \text{LayerNorm}\big(\text{x} + \text{MHA}(\text{x})\big),
\end{align*}
where $\attn$ is a dot product attention head, and $\text{x}$ is the input sequence.
The output of the MHA layer is then fed into the FFN layer, which consists of $N$ filters:
\vspace{-1mm}
\begin{align*}
\text{FFN}(\text{x}) = \big(\sum_{i=1}^{N} \textrm{W}^{\textsc{(2)}}_{:,i} \sigma (\textrm{W}^{\textsc{(1)}}_{i,:}\text{x} + b^{\textsc{(1)}}_i)\big) + b^{\textsc{(2)}}, \quad
\text{x}_\mathrm{{out}} = \text{LayerNorm}\big(\text{x}_\textsc{{mha}} + \text{FFN}(\text{x}_\textsc{{mha}})\big),
\end{align*}
where $\textrm{W}^{\textsc{(1)}}, \textrm{W}^{\textsc{(2)}}, b^{\textsc{(1)}}$ and $b^{\textsc{(2)}}$ are the FFN parameters, and $\sigma$ is the activation function, typically GELU~\cite{hendrycks2016gaussian}.
Note that ($H$, $N$) is (12, 3072) for BERT$_\textsc{base}$, and (16, 4096) for BERT$_\textsc{large}$.
We also denote $L$ as the number of Transformer layers (e.g., 12 for BERT$_\textsc{base}$).

\textbf{Granularity of Pruning.}
Our framework considers the structured pruning of both heads in MHA and filters in FFN layers.
We do not prune the embedding and the final classifier, as computation of those layers takes a negligible portion of the total inference latency.
Since our pruning framework always produces a smaller dense architecture, the model can be readily accelerated without the need of specialized hardware logic, which is often required for unstructured sparsity to gain latency speedup.

\textbf{Notations.}
We pose the pruning problem as finding a sparse mask for the heads and filters.
To formalize this, we introduce mask variables associated with the outputs of heads and filters:
\vspace{-1.5mm}
\begin{gather*}
\small
\text{MHA}(\text{x}; \text{m}_l^{\textsc{mha}}) = \sum_{i=1}^{H} m_{l,i}^{\textsc{mha}} \circ \attn_{i}(\text{x}), \\[-3mm]
\text{FFN}(\text{x}; \text{m}_l^{\textsc{ffn}}) = \big(\sum_{i=1}^{N} m_{l,i}^{\textsc{ffn}} \circ \textrm{W}^{\textsc{(2)}}_{:,i} \sigma (\textrm{W}^{\textsc{(1)}}_{i,:}\text{x} + b^{\textsc{(1)}}_i) \big) +b^{\textsc{(2)}},
\end{gather*}
where $\text{m}_l^{\textsc{mha}} \hspace{-0.5mm} \in \hspace{-0.5mm} \mathbb{R}^{H}$
and $\text{m}_l^{\textsc{ffn}} \hspace{-0.5mm} \in \hspace{-0.5mm} \mathbb{R}^{N}$ are the mask variables for MHA and FFN in the $l$-th layer, respectively, and $m_{l,i}^{\textsc{mha}}$  and $m_{l,i}^{\textsc{ffn}}$ are their $i$-th elements.
Furthermore, $\circ$ denotes the Hadamard product.
Originally, the mask variables are all initialized to 1, which does not change the model outputs.
After pruning, the mask variables become zero or any nonzero values, affecting the model accuracy and sparsity.
Especially, setting $m_{l,i}^{\textsc{mha}}$ and $m_{l,i}^{\textsc{mha}}$ as zero is equivalent to pruning the $i$-th head and filter, respectively.

Overall, there are $LH$ head mask variables and $LN$ filter mask variables, summing up to $L(H+N)$ number of total mask variables in a Transformer model.
To simplify notations, we additionally define
$\text{m}^\textsc{mha} \in \mathbb{R}^{LH}$, 
$\text{m}^\textsc{ffn} \in \mathbb{R}^{LN}$, and
$\text{m} \in \mathbb{R}^{L(H+N)}$
as flattened vectors of the head, filter, and total mask variables, respectively, across all layers.
In what follows, we discuss how to find the optimal sparse masks under a given cost constraint and how to adjust their values to recover accuracy.


\begin{figure}[!t]
\centering
{
    \includegraphics[width=0.75\textwidth]{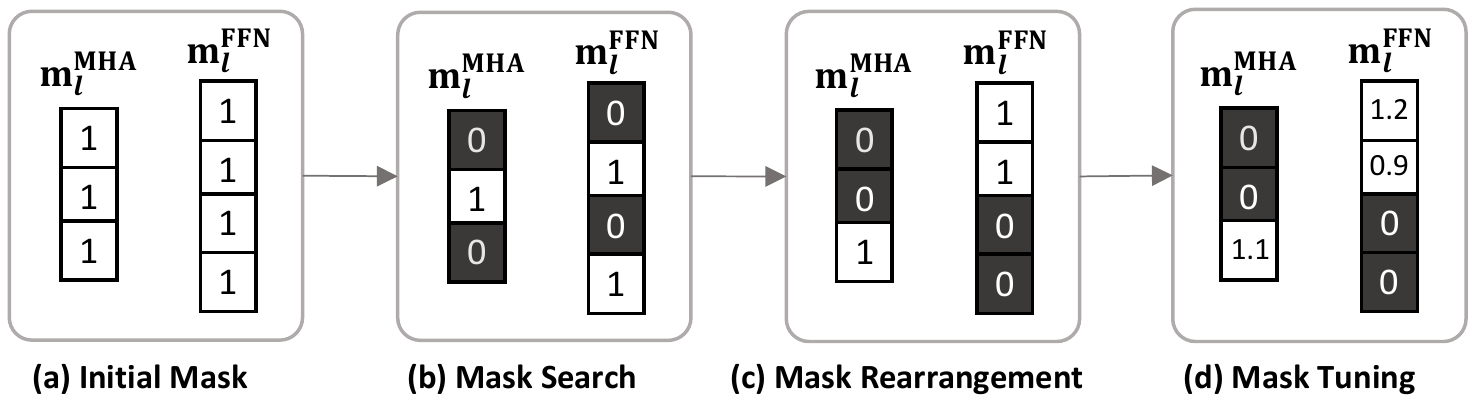}
    }
\caption{
Overview of our pruning framework.
(a) The mask variables 
are initialized as 1.
Then they undergo the three-stage pipeline of (b) mask search (\sref{subsec:search}), (c) rearrangement (\sref{subsec:rearrange}), and (d) tuning (\sref{subsec:tuning}).
}
\vspace{-4mm}
\label{fig:procedure}
\end{figure}

\vspace{-1mm}
\subsection{Framework Overview}
\label{subsec:overview}

\fref{fig:overview}(b) and~\fref{fig:procedure} illustrate the overview of our framework.

\textbf{Inputs.}
Our framework has 3 inputs: a Transformer model; a sample dataset; and a resource constraint.
The input Transformer model should contain weights fine-tuned for a downstream task.
The sample dataset is a small partition of the training dataset (typically 1--2K examples) for the downstream task.
The resource constraint can be given either as the number of floating point operations (FLOPs) or as an actual latency on target hardware.
In the later case, we further assume that a latency lookup table for the target hardware is provided.

\textbf{Compression Pipeline.}
As illustrated in \fref{fig:procedure},
our framework consists of 3 stages: Fisher-based mask search; Fisher-based mask rearrangement; and mask tuning.
During the \textit{Fisher-based mask search }stage (\sref{subsec:search}), we search for a binary mask applied to the heads and filters based on the Fisher information of the mask variables.
Intuitively, the mask variables with relatively higher Fisher information are considered more important, and they should be less likely to be pruned~\cite{lecun1990optimal,liu2021group,molchanov2019importance}.
As finding the optimal mask that minimizes the Fisher information loss is intractable due to the large size of the full Fisher matrix,
we propose a lightweight search algorithm that finds the optimal mask under reasonable approximations.
Second, in the \textit{Fisher-based mask rearrangement} stage (\sref{subsec:rearrange}), the framework adjusts the searched mask patterns to better take into account the intra-layer interactions between the mask variables. 
Lastly, in the \textit{mask tuning} stage (\sref{subsec:tuning}), the framework tunes the nonzero mask variables to recover the accuracy drop by reconstructing the layer-wise output signal.

\section{Methodology}
\label{sec:methodology}

The pruning problem can be seen as finding an optimal mask under a sparsity constraint.
However, without retraining, the problem becomes intractable.
To address this, we decompose Transformer pruning into three sub-problems, each of which can be efficiently solved; the first two stages of our pipeline address the problems of finding an optimal \textit{binary} mask, and the last stage further optimizes it into a \textit{real-valued} mask.

Note that the number of the mask variables is much less than the number of the parameters in a Transformer (e.g., 37K vs. 110M in case of BERT$_\textsc{base}$).
This allows the framework to use only a small number of examples without overfitting to the sample dataset, and thus to be extremely faster than the retraining-based pruning methods which typically use the entire dataset.
As the framework keeps the model ``as is'' and only decides the mask variables, we henceforth regard the model parameters as constants and consider the mask variables as the only parameters for our pruning problem.

\textbf{Problem Formulation.}
We formulate Transformer pruning as a constrained optimization problem on the mask $\text{m}$:
\begin{align}
    \argmin_{\text{m}} \loss(\text{m}) \label{eq:original_objective} \quad \text{s.t.} \quad \text{Cost}(\text{m}) \leq C
\end{align}
where $\loss$ denotes the loss function, $\text{Cost}$ is the FLOPs/latency of the architecture pruned by the mask, and $C$ is the given FLOPs/latency constraint.
Unfortunately, such a problem is generally intractable as $\text{Cost}$ is usually a function of $l_0$-norm of the mask $\text{m}$, which is non-differentiable.
Thus, in what follows, we introduce several assumptions and approximations to simplify the~problem.

We start by approximating the loss function using the second-order Taylor expansion around the initial mask  $\mathbb{1}$:
\vspace{-2mm}
\begin{align}
    \loss(\text{m}) & \approx \loss(\mathbb{1}) - \text{g}^\intercal (\mathbb{1} - \text{m}) + \frac{1}{2}(\mathbb{1} - \text{m})^\intercal \text{H} (\mathbb{1} - \text{m}) \\
    &\approx \loss(\mathbb{1}) + \frac{1}{2}(\mathbb{1} - \text{m})^\intercal \text{H} (\mathbb{1} - \text{m}) , \label{eq:lm_nograd}
\end{align}
where $\text{g} = \mathbb{E}[\frac{\partial}{\partial \text{m}} \loss(\mathbb{1})]$ and $\text{H} = \mathbb{E}[\frac{\partial^2}{\partial \text{m}^2} \loss(\mathbb{1})]$.
\eref{eq:lm_nograd} is deduced from an assumption that the model has converged to a local minima, where the gradient term is close to 0~\cite{lecun1990optimal}.
As $\loss(\mathbb{1})$ is a constant, we can rewrite the optimization objective as follows:
\begin{align}
    \argmin_{\text{m}} \loss(\text{m}) 
    \approx \argmin_{\text{m}} (\mathbb{1} - \text{m})^\intercal \text{H} (\mathbb{1} - \text{m}). \label{eq:argmin_hessian}
\end{align}
\eref{eq:argmin_hessian} shows that the optimal mask is determined by the Hessian of the loss with respect to the mask variables.
Since forming the exact Hessian matrix explicitly is infeasible, we approximate the Hessian $\text{H}$ with the (empirical) Fisher information matrix $\mathcal{I}$ of the mask~variables:
\begin{align}
\small
    \mathcal{I} := \frac{1}{|\mathcal{D}|} \sum_{(x, y) \in \mathcal{D}} \big(\frac{\partial}{\partial \text{m}}\loss(x, y; \mathbb{1})\big) \big(\frac{\partial}{\partial \text{m}}\loss(x, y; \mathbb{1})\big)^\intercal, \label{eq:fim}
\end{align}
where $\mathcal{D}$ is the sample dataset and $(x, y)$ is a tuple of an input example and its label.

\subsection{Fisher-based Mask Search}
\label{subsec:search}

\textbf{Diagonal Approximation of the Fisher Information Matrix.}
It is intractable to solve the optimization objective in~\eref{eq:argmin_hessian} using the full Fisher information matrix $\mathcal{I}$.
Thus, we first make a simple assumption that $\mathcal{I}$ is \textit{diagonal}.
This further simplifies \eref{eq:argmin_hessian} as follows:
\begin{gather}
    \argmin_{\text{m}} \loss(\text{m}) \approx \argmin_{\text{m}} \sum_i (1 - m_i)^2 \mathcal{I}_{ii}, \label{eq:argmin_importance}
\end{gather}
Since we restrict the possible mask values to either 0 or 1, the following can be derived from \eref{eq:argmin_importance}:
\begin{gather}
    \argmin_{\text{m}} \loss(\text{m}) \approx \argmin_{\text{m}} \sum_{i \in Z(\text{m})} \mathcal{I}_{ii} \label{eq:sum_importance} \quad
    \text{where} \quad Z(\text{m}) := \{i \ | \ m_i = 0 \}.
\end{gather}
We can interpret each diagonal element of $\mathcal{I}$ as the \textit{importance score} of the head/filter associated with the mask variable,
and \eref{eq:sum_importance} as a process of minimizing the total importance scores of the pruned heads and filters.
Such an importance score has also been introduced in~\cite{molchanov2019importance, theis2018faster} to guide pruning.

\setlength{\textfloatsep}{6pt}
\begin{algorithm}[tb]
   \caption{Mask Search with a FLOPs Constraint}
   \label{alg:search_flops}
   
    {\bfseries Input:} FLOPs constraint $C$, diagonal Fisher information matrix $\mathcal{I}$
   
\begin{algorithmic}[1]
    \STATE \textbf{for} {$n=0$ {\bfseries to} $LH$} \textbf{do} \label{alg:search_flops:for} 
    \hspace*{\fill}{$\triangleright$ \# remaining heads}
        \STATE \quad  $k_1 = LH - n$ \hspace*{\fill}{$\triangleright$ \# heads to prune}
        \STATE \quad $\text{HI} = $ indicies of $k_1$ least important heads
        \STATE \quad $f = \lfloor (C - n \text{F}_{\text{head}}) / \text{F}_\text{filter} \rfloor $ \label{alg:search_flops:num_filters}
        \hspace*{\fill}{$\triangleright$ \# remaining filters}
        \STATE \quad $k_2 = LN - f$
        \hspace*{\fill}{$\triangleright$ \# filters to prune}
        \STATE \quad $\text{FI} = $ indicies of $k_2$ least important filters
        \STATE \quad $S[n] = \sum_{i \in \text{HI} \cup \text{FI}} \mathcal{I}_{ii}$
        \STATE \quad $R[n] = (\text{HI}, \text{FI})$
    \STATE \textbf{end for}
    \STATE $n^\ast = \argmin_n S[n]$
    \hspace*{\fill}{$\triangleright$ optimal \# remaining heads} \label{alg:search_flops:argmin_s}
    \STATE $\text{HI}^\ast, \text{FI}^\ast = R[n^\ast]$ \label{alg:search_flops:r_n}
    \hspace*{\fill}{$\triangleright$ indicies of heads/filters to prune}
    \STATE Initialize $\text{m}^{\textsc{MHA}}$ and $\text{m}^{\textsc{FFN}}$ as $\mathbb{1}$
    \STATE $\text{m}^{\textsc{MHA}}[\text{HI}^\ast] = 0$ \hspace*{\fill}{$\triangleright$ prune the selected heads}
    \STATE $\text{m}^{\textsc{FFN}}[\text{FI}^\ast] = 0$ \hspace*{\fill}{$\triangleright$ prune the selected filters}
\end{algorithmic}

     {\bfseries Output:} $\text{m}^\ast = (\text{m}^{\textrm{MHA}}$, $\text{m}^{\textrm{FFN}})$
\end{algorithm}

\textbf{Solving FLOPs-constrained Problem.}
We need to solve~\eref{eq:sum_importance} given a cost constraint.
For a given target FLOPs cost, denoted by C, we can formulate the binary mask search problem as follows:
\begin{gather}
    \argmin_{\text{m}} \sum_{i \in Z(\text{m})} \mathcal{I}_{ii} \quad
    \text{s.t.} \quad \text{F}_{\text{head}} || \text{m}^{\textsc{mha}} ||_0 + \text{F}_{\text{filter}} || \text{m}^{\textsc{ffn}} ||_0 \leq C \label{eq:flops_objective},
\end{gather}
where $\text{F}_\text{head} \in \mathbb{R}$ and $\text{F}_\text{filter} \in \mathbb{R}$  are the FLOPs for computing a head and a filter, respectively.
Note that the number of FLOPs of a head/filter is constant across all layers.
While such an optimization problem can be generally solved by a knapsack algorithm~\cite{aflalo2020knapsack, shen2021halp}, the following observations allow a faster polynomial-time solution:
(1) having more heads and filters unpruned always optimizes \eref{eq:flops_objective} since the diagonal elements of $\mathcal{I}$ are non-negative; and
(2) if a certain number of heads needs to be pruned, they should be the ones with the lowest importance scores because each head accounts for the same amount of FLOPs.
The same statement also holds for pruning filters.
The two observations lead to our mask search algorithm described in Algorithm~\ref{alg:search_flops}.

Algorithm~\ref{alg:search_flops} partitions the solution space by the total number of remaining heads in the pruned architecture ($n$ in line~\ref{alg:search_flops:for}).
For each $n$, the number of remaining filters should be the largest possible number that satisfies the cost constraint by observation (1), which can be described as $f$ in line~\ref{alg:search_flops:num_filters}.
Then by observation (2), the heads/filters with the lowest important scores are selected to be pruned. 
Therefore, $S[n]$ is a solution of~\eref{eq:flops_objective} under additional constraint of fixing the number of remaining heads to be $n$.
When the loop terminates, 
the output is the mask that minimizes $S[n]$ across all possible $n$ (line~\ref{alg:search_flops:argmin_s} and \ref{alg:search_flops:r_n}).
In~\sref{appendix:optimality_proof}, we prove that
the output mask $\text{m}^\ast$ of Algorithm~\ref{alg:search_flops} is optimal.
That is, any other mask $\text{m}$ satisfying the given FLOPs constraint will result in a higher loss:
\begin{equation}
\label{eqn:from_theorem:optimal_mask}
   \sum_{i \in Z(\text{m}^\ast)} \mathcal{I}_{ii} \leq \sum_{i \in Z(\text{m})} \mathcal{I}_{ii}.
\end{equation}

\textbf{Solving Latency-constrained Problem.}
If the cost constraint is given in terms of latency on target hardware, 
we have a new optimization problem with a different cost constraint than \eref{eq:flops_objective}:
\begin{gather}
    \argmin_{\text{m}} \sum_{i \in Z(\text{m})} \mathcal{I} \label{eq:latency_objective} \quad
    \text{s.t.} \quad \sum_{l=1}^L \text{LAT}(\text{m}^{\textsc{mha}}_l) + \sum_{l=1}^L \text{LAT}(\text{m}^{\textsc{ffn}}_l) \leq C,
\end{gather}
where the function $\text{LAT}$ indicates the latency of a MHA/FFN layer after pruning.
We assume that a latency lookup table on the target hardware is provided so that evaluating $\text{LAT}$ takes negligible time.

Unfortunately, the latency constraint makes the problem more challenging 
as directly applying Algorithm~\ref{alg:search_flops} is no longer possible.
This is because $\text{LAT}$ is \textit{not} linear to the number of remaining heads or filters after pruning~\cite{radu2019performance}, as shown in \fref{fig:lat} (Left).
We can interpret this as follows:
(1) with a sufficient number of heads/filters in a layer, the hardware resources such as parallel cores can be fully utilized, resulting in latency roughly proportional to the number of heads/filters; and
(2) otherwise, the hardware is underutilized and a constant overhead dominates the latency~\cite{kwon2020nimble, ma2020rammer}.
Thus, pruning more heads/filters below a certain threshold does not translate into actual speedup.

\begin{table}[!t]
	\begin{minipage}{0.62\linewidth}
    \vspace{8mm}
		\begin{center}
    \includegraphics[width=1\columnwidth]{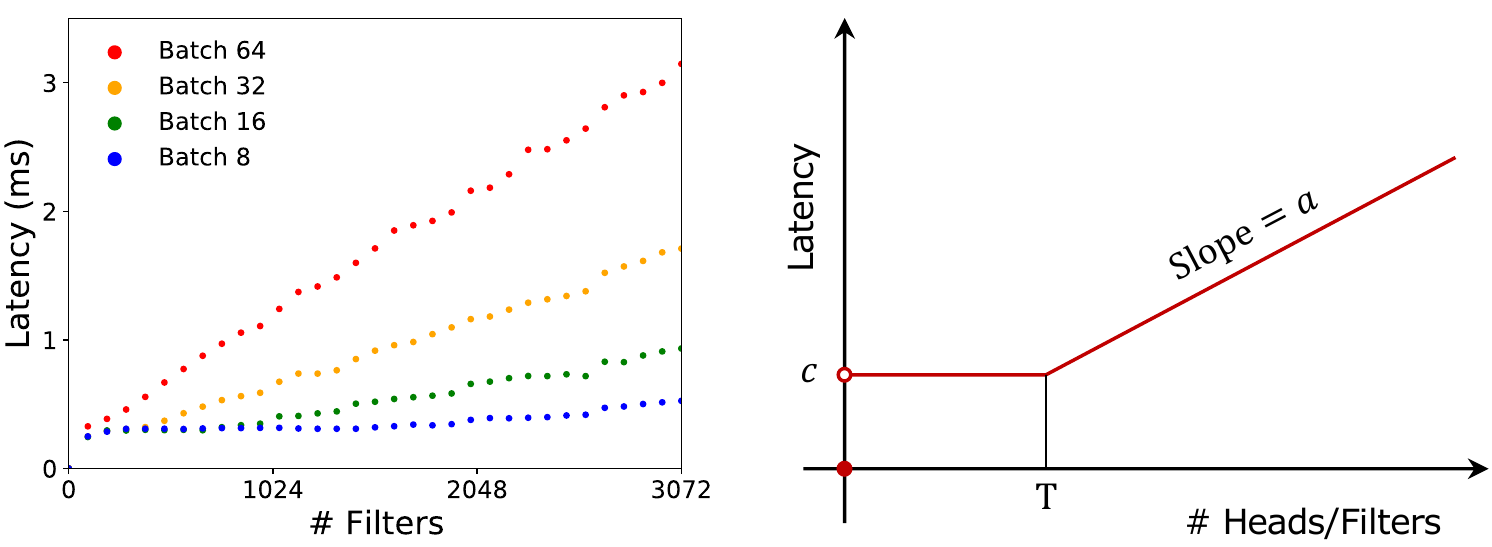}
  \end{center}
    \captionof{figure}{
    (Left) Real latency of a single FFN layer with different numbers of remaining filters. (Right) Schematic plot for the approximated latency as a piece-wise linear function.}
    \label{fig:lat}
	\end{minipage}
	\hfill
	\begin{minipage}{0.33\linewidth}
  \begin{center}
    \includegraphics[width=1\columnwidth]{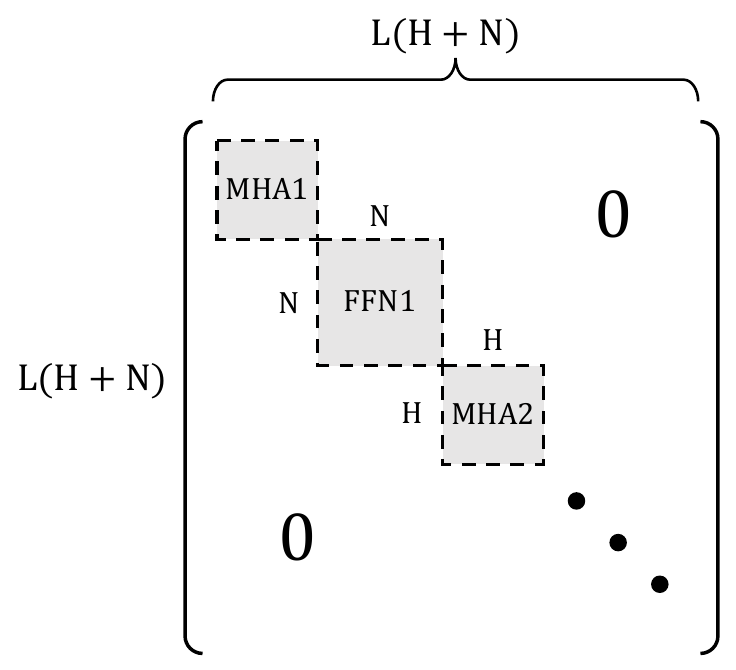}
  \end{center}
  \captionof{figure}{
    Illustration of the block diagonal Fisher matrix.
    }
\label{fig:block-diagonal}
	\end{minipage}
	\vspace{-3mm}
\end{table}

Based on the above analysis, we approximate $\text{LAT}$ as a piece-wise linear function as in \fref{fig:lat} (Right)
such that $\text{LAT}(\text{m}_l)$ is 0 if $||\text{m}_l||_0 = 0$,
$c$ if $0 < ||\text{m}_l||_0 \leq T$, and 
$a(||\text{m}_l||_0 - T) + c \ $ if $||\text{m}_l||_0 > T$,
where $c \in \mathbb{R}$ is the constant overhead, $T \in \mathbb{N}$ is the threshold number of heads/filters that the latency starts to become linear, and $a \in \mathbb{R}$ is the slope of the linear part.
This can be easily obtained by fitting the actual latency data in the lookup table with the minimum mean squared error.

The piece-wise linear approximation of $\text{LAT}$ allows us to extend Algorithm~\ref{alg:search_flops} to the setting with latency constraints.
The core idea is to separately consider the constant part and the linear part of $\text{LAT}$; after handling the constant part, we can apply the Algorithm~\ref{alg:search_flops} to the linear part.
The detailed modification to Algorithm~\ref{alg:search_flops} is described in~\sref{appendix:latency_algorithm}.

\vspace{-2mm}
\subsection{Fisher-based Mask Rearrangement}
\label{subsec:rearrange}
\vspace{-1mm}

\textbf{Block Diagonal Approximation of the Fisher Information Matrix.}
Although it simplifies the problem, the diagonal assumption in \sref{subsec:search} alone might not find the best solution, as it does not take into account the interactions between different mask variables.
For example, if there are two attention heads playing a similar role in a layer, pruning only one of them might not affect the model accuracy.
However, when both of them are pruned, the model accuracy can be significantly degraded.
Such interactions are captured by the non-diagonal elements of the Fisher information matrix, which were ignored in the previous stage.
Thus, we can better consider the interactions in our pruning problem by using a \textit{block diagonal} approximation to the Fisher matrix, where a block corresponds to a MHA layer or a FFN layer as illustrated in~\fref{fig:block-diagonal}.

However, the block diagonal approximation results in an intractable optimization problem over the binary mask.
To alleviate this, we use the results from the previous stage to \textit{warm start} the optimization problem.
First, we constrain the number of heads/filters to prune for each layer to be the same as the binary mask we obtained in the first stage.
In other words, given the mask $\text{m}^\ast$ obtained in \sref{subsec:search}, we constrain 
$|| \text{m}_l ||_0$ to be equal to  $||\text{m}^{\ast}_l ||_0 $ for each layer $l$.
Second, we use the mask $\text{m}^\ast$ as the starting point of the greedy search to solve the new optimization problem. 

Given the two assumptions that (i) there is no interaction between the mask variables in different layers (i.e., the block diagonal approximation), and (ii) the number of heads/filters to prune are pre-determined for each layer (i.e., warm-start), \eref{eq:argmin_hessian} breaks down to a set of \textit{layer-wise} optimization problems, as follows based on the derivation in~\sref{appendix:derivation_block_diagonal}:
\begin{gather}
    {\hat{\text{m}}}_l = \argmin_{\text{m}_l} (\mathbb{1} - \text{m}_l)^\intercal \mathcal{I}_l (\mathbb{1} - \text{m}_l), \label{eq:layerwise_opt}
\end{gather}
where $\mathcal{I}_l$ is the $l$-th diagonal block of $\mathcal{I}$.
We approximately solve this problem with a greedy algorithm.
After initializing the mask $\text{m}_l$ as $\text{m}^{\ast}_l$ (i.e., warm-start),
we pick for every round a pruned head (or filter) with the highest Fisher information and exchange it with an unpruned head (or filter) in the current mask if that can further optimize \eref{eq:layerwise_opt}.
After every pruned head/filter goes through one round, we obtain an approximate solution to \eref{eq:layerwise_opt}.

Because this process does not change the number of heads/filters in each layer, the obtained mask $\hat{\text{m}}_l$ results in the same FLOPs/latency as that of the mask $\text{m}^{\ast}_l$ searched in~\sref{subsec:search}.
In effect, this process \textit{rearranges} the binary mask variables of each layer to find a better arrangement of pruning locations and capture the intra-layer interactions. 

\subsection{Mask Tuning}
\label{subsec:tuning}
\vspace{-1mm}

In the previous two stages, the possible mask values are restricted to either 0 or 1 in order to simplify the search process.
In this stage, we further relax this restriction.
The \textit{nonzero} variables in the mask $\hat{\text{m}}$ from \sref{subsec:rearrange} are tuned to any real values such that the pruned model recovers its accuracy.

\textbf{Layer-wise Reconstruction via Linear Least Squares.}
We tune the mask variables toward minimizing the \textit{layer-wise reconstruction error}, similarly to~\cite{he2017channel}.
From the first to the last layer, we reconstruct the output activation of the original model with the remaining heads/filters in the pruned model.
This can be formally written as follows:
\vspace{-1mm}
\begin{gather}
    \argmin_{\text{m}_l} || \text{x}  + \mathrm{layer}(\text{x}; \text{m}_l) - \big(\text{x}' + \mathrm{layer}(\text{x}'; \mathbb{1})\big) ||_2^2, \label{eq:reconstruction}
\end{gather}
where $\mathrm{layer}$ is either MHA or FFN, and $\text{x}$ and $\text{x}'$ are the inputs to the layer of the pruned model and the original model, respectively.
Here we compare the activations after the residual connection.
Note that this stage does not incur any change in model FLOPs/latency, as we only tune the nonzero mask variables.
We show in~\sref{appendix:derivation_lsa} that \eref{eq:reconstruction} can be reduced to a \textit{linear least squares} problem of
$\argmin_{\text{m}^{}_l} || \text{A} \text{m}^{}_l - \text{b}  ||_2^2$, where the matrix $\text{A}$ denotes head/filter-wise output activations of the model pruned by the binary mask and the vector $\text{b}$ is the difference between the output activations of the two models.
Concretely, when there are $T$ tokens in the sample dataset and $D$ is the hidden size of the model, the size of the matrix $\text{A}$ is $TD \times H$ for head masks and $TD \times N$ for filter masks.

Due to the large size of the matrix $\text{A}$, naively solving the least squares problem can lead to numerically unstable results.
To address this, our framework uses the LSMR solver in CuPy~\cite{nishino2017cupy} with a regularization hyperparameter (i.e., \texttt{damp}).
Concretely, we re-parameterize the least squares problem as $\argmin_{\text{r}^{}_l} || \text{A} \text{r}^{}_l + \text{A} \cdot \mathbb{1} - \text{b}  ||_2^2$ where $\text{m}_l =  \mathbb{1} + \text{r}_l$, and solve it with the \texttt{damp} value fixed to 1.
Then, to prevent the case in which the tuned mask rather hurts the accuracy, we restrict the acceptable range of the tuned mask variables to [-10, 10].
When the solver finds a layer mask that exceeds this range, we discard the mask for that layer and stop mask tuning.
In our experiments, we find that the aforementioned heuristics make the mask tuning process highly stable across different models, tasks, and seeds. 
Furthermore, while the use of the heuristics involves the two hyperparameters (i.e., \texttt{damp} and the acceptable range), we empirically find that these need not be tuned for different tasks and models.
In all of our experiments, we fixed the two hyperparameter values as we mentioned here.

\begin{figure*}[!t]
\centering
\includegraphics[width=\textwidth]{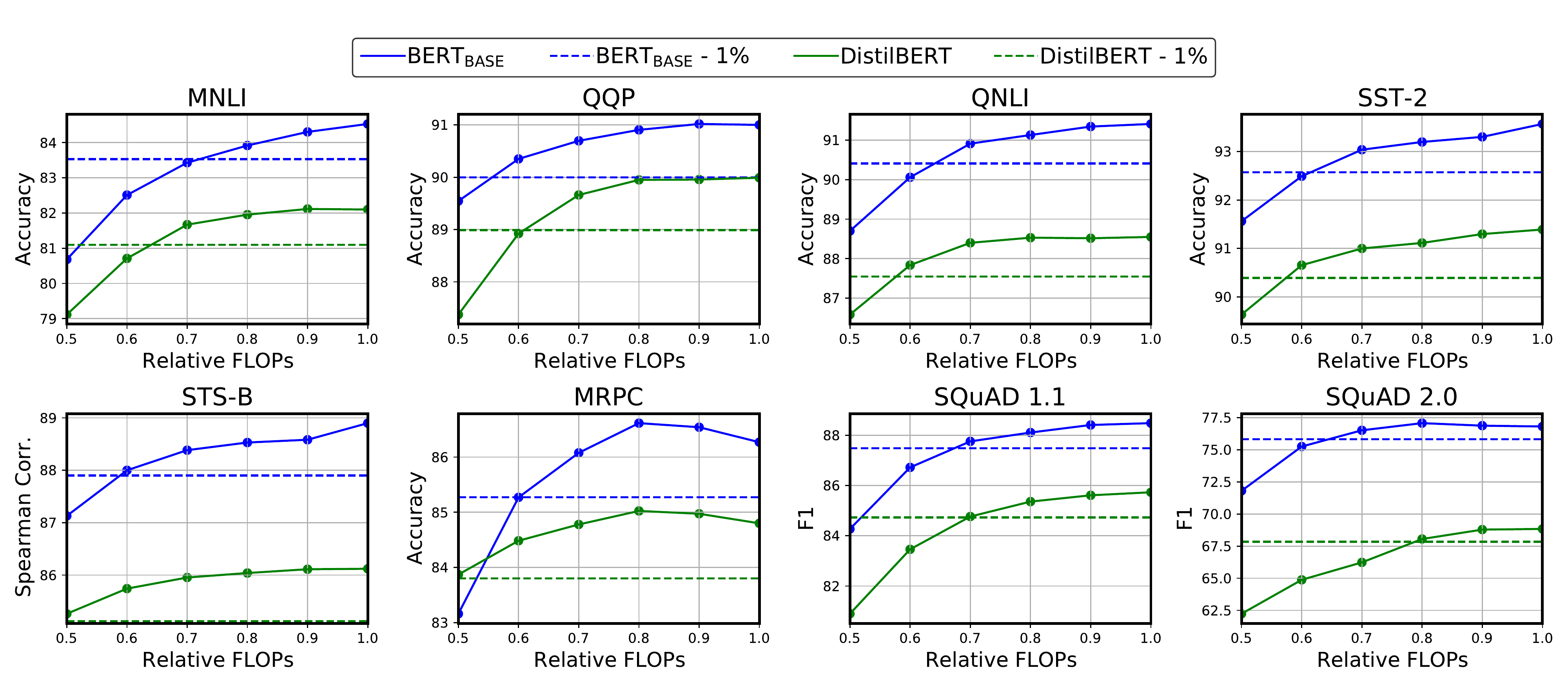}
\vspace{-7mm}
\caption{
Accuracy of our pruning method applied to BERT$_\textsc{BASE}$ and DistilBERT with different FLOPs constraints. 
The dashed horizontal lines indicate 1\% accuracy drop from the baseline models.
Note that these results can be achieved in only 39 and 135 seconds for GLUE and SQuAD benchmarks, respectively, on a single GPU system, as described in \tref{tab:time_breakdown} (in~\sref{appendix:time_breakdown}).
}
\vspace{-3mm}
\label{fig:flops_result}
\end{figure*}

\begin{table*}[!t]
\caption{
Latency speedup of BERT$_\textsc{base}$ on a single NVIDIA V100 GPU with different batch sizes.
The latency is measured using PyTorch.
We constrain the accuracy degradation to be at most 1\% from the baseline accuracy,
and we report the largest speedup among those that satisfy the constraint.
}\label{tab:latency_result}
\vspace{-2mm}
\centerline{
    \centering
    \small{
    \setlength{\tabcolsep}{4.5pt}{
      \begin{tabular}{c|cccccccc|c}
        \toprule
        Batch size & {MNLI} & {QQP}  & {QNLI} & {SST-2} & {STS-B} & {MRPC} & {SQuAD$_{1.1}$} & {SQuAD$_{2.0}$} & Geo. mean\\ 
        \midrule         
        32 &  1.27$\times$ & 1.42$\times$ & 1.42$\times$ & 1.23$\times$ & 1.34$\times$ & 1.36$\times$ & 1.33$\times$ & 1.37$\times$ & 1.34$\times$ \\
        256 & 1.34$\times$ &	1.54$\times$ &	1.53$\times$ &	1.56$\times$ &	1.54$\times$ &	1.55$\times$ &	1.34$\times$ &	1.40$\times$ & 1.47$\times$\\
        \bottomrule
        \end{tabular} 
    }
    }
}
\end{table*}

\section{Evaluation}
\label{sec:eval}
\vspace{-1mm}

\subsection{Experimental Setup}
\label{subsection:experimental_setup}

Our framework is implemented on top of PyTorch~\cite{paszke2019pytorch} and the HuggingFace Transformers~\cite{wolf2020transformers} library.
We evaluate the effectiveness of our approach using BERT$_\textsc{BASE}$~\cite{devlin2018bert} and DistilBERT~\cite{sanh2019distilbert} on GLUE~\cite{wang2018glue} and SQuAD~\cite{rajpurkar2018know, rajpurkar2016squad} benchmarks.
We use 2K examples from the training sets for pruning, and we evaluate the resulting models on the development sets.
All of the results are averaged over the runs with 10 different seeds.
More details on the experimental setup can be found in \sref{appendix:eval_details}.

\subsection{Performance Evaluation}
\label{subsection:performance}

\textbf{FLOPs.} \fref{fig:flops_result} shows the accuracy of BERT$_\textsc{BASE}$ and DistilBERT with different
FLOPs constraints on GLUE and SQuAD datasets.
As can be seen in the plots, with only 1\% of accuracy drop, BERT$_\textsc{BASE}$ achieves 60--70\% of the original FLOPs  for all tasks. 
DistilBERT also shows a similar pattern and achieves up to 50\% FLOPs reduction (in STS-B and MRPC) even though it is already a compressed architecture.
More results using larger sample datasets are provided in~\sref{appendix:sample_dataset}.

\textbf{Latency.}
We further measure the latency on real hardware by pruning BERT$_\textsc{BASE}$ with latency constraints and deploying the resulting models on an NVIDIA V100 GPU.
\tref{tab:latency_result} lists the latency speedup with maximum accuracy drop of 1\% for GLUE and SQuAD datasets.
With batch size of 256, we achieve speedup of 1.47$\times$ on average and up to 1.56$\times$.

\begin{figure*}[!t]
\centering
\includegraphics[width=1\textwidth]{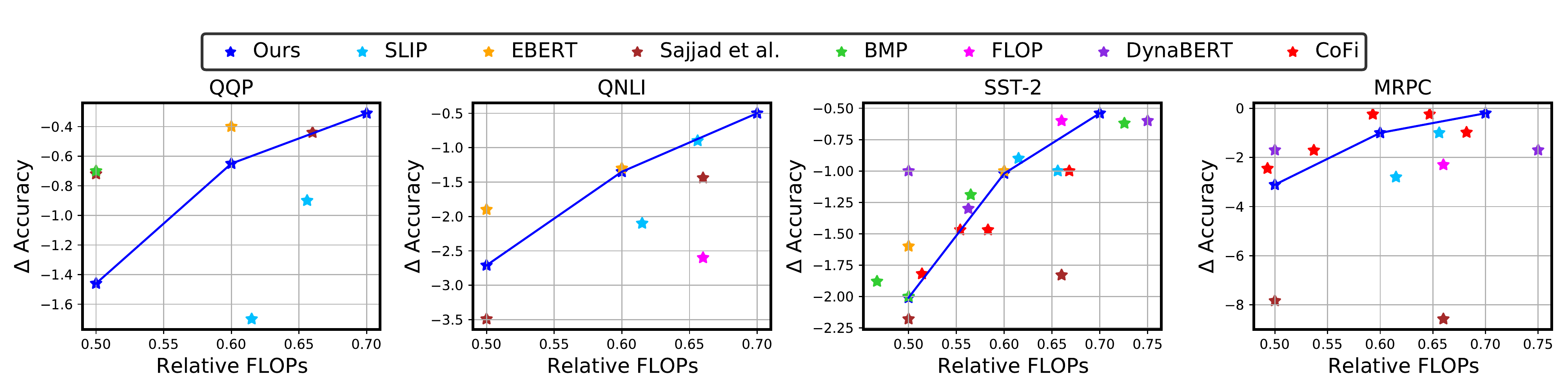}
\caption{
Amount of accuracy degradation from the baseline when pruning BERT$_\textsc{BASE}$ using our method and the prior structured pruning methods with different relative FLOPs. 
Note that our method does not require retraining, whereas all the other methods involve significant retraining overheads as described in \tref{tab:pruning_cost}.
}
\label{fig:comparison}
\end{figure*}

\subsection{Comparison with the Prior Methods}
\label{subsection:comparison}

\textbf{FLOPs and Accuracy Comparison.}
Here, we compare our method with the prior structured pruning methods for Transformers including 
Flop~\cite{wang2019structured}, 
SLIP~\cite{lin2020pruning}, 
Sajjad et al. \cite{sajjad2020effect}, 
DynaBERT~\cite{hou2020dynabert}, 
EBERT~\cite{liu2021ebert}, 
Block Movement Pruning (BMP)~\cite{lagunas2021block}, and CoFi~\cite{xia2022structured} by the FLOPs-accuracy trade-off of BERT$_\textsc{BASE}$ on GLUE tasks.
We use the results \textit{without} knowledge distillation and data augmentation reported in each paper. 
Since the baseline accuracy differs slightly from paper to paper, 
we compare the amount of the accuracy drop from the baseline instead of the absolute accuracy.
The results are plotted as~\fref{fig:comparison}.
We include the comparison details and full table in~\sref{appendix:comparison_details}.

Interestingly, our method exhibits comparable or better results than the prior methods \textit{without} any model retraining and with substantially lower pruning costs.
This empirically demonstrates that retraining and a complex pruning pipeline are \textit{not} necessary for moderate level of pruning of Transformers.
For high sparsity, we find that our framework \textit{with} retraining works comparably to or better than the prior methods at the same pruning cost (See~\sref{appendix:high_sparsity}).

\begin{table}[!t]
	\begin{minipage}{0.47\linewidth}
	\vspace{-6mm}
		\centering
		\footnotesize
        \setlength{\tabcolsep}{3pt}
\caption{
Pruning cost comparison between the prior structured pruning methods and ours.
We compare the number of training epochs and the end-to-end (E2E) time required for pruning.
}
\vspace{2mm}
\label{tab:pruning_cost}
    \centering
    \small{
    \setlength{\tabcolsep}{5pt}{
       \begin{tabular}{lccc}
        \toprule
        & \# Epochs & E2E time (hr) \\ 
        \midrule
        DynaBERT~\cite{hou2020dynabert} & 4 & 12 \\ 
        EBERT~\cite{liu2021ebert} & 6 & 5 \\ 
        BMP~\cite{lagunas2021block} & 20 & 17 \\ 
        CoFi~\cite{xia2022structured} & 40 & 33 \\
        \midrule
        \ha Ours & \textbf{0} & \textbf{0.01} \\ 
        \bottomrule
        \end{tabular} 
    }
    }
	\end{minipage}
	\hfill
	\begin{minipage}{0.52\linewidth}
  \begin{center}
    \includegraphics[width=1\columnwidth]{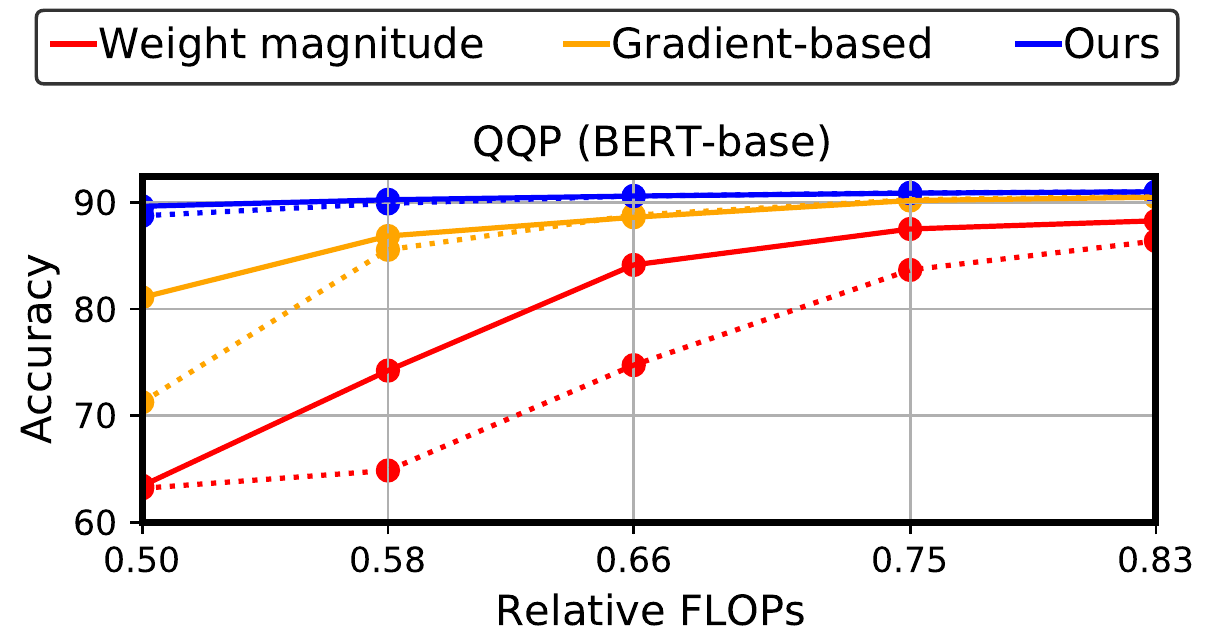}
  \end{center}
  \vspace{-2mm}
  \captionof{figure}{
Retraining-free accuracy without (dotted) and with (solid) mask tuning.
}
\label{fig:effectiveness}
	\end{minipage}
	\vspace{-6mm}
\end{table}

\textbf{Retraining Cost.}
We select DynaBERT~\cite{hou2020dynabert}, EBERT~\cite{liu2021ebert}, BMP~\cite{lagunas2021block}, and CoFi~\cite{xia2022structured} that achieve comparably good accuracy in \fref{fig:comparison}, and we systematically analyze their end-to-end retraining costs on MNLI dataset.
As shown in \tref{tab:pruning_cost}, these methods require 5$-$33 hours of retraining.
On the other hand, our method finishes in less than a minute, which is 2$-$3 orders of magnitude faster. 
We also highlight that this training latency analysis only accounts for a \textit{single} hyperparameter, and the entire cost should be multiplied by the size of the hyperparameter space.
While the prior methods rely on a considerable number of hyperparameters, ours introduce only two hyperparameters (in~\sref{subsec:tuning}) which we fix for all experiments.
See \sref{appendix:retraining_cost} for more details.

\begin{table*}[!t]
\caption{
Ablation of our mask search, rearrangement, and tuning methods, described in \sref{sec:methodology}.
We use BERT$_\textsc{BASE}$ as the baseline model, and we prune it with a 60\% FLOPs constraint.
}
\vspace{-2mm}
\label{tab:ablation}
    \centering
    \small{
    \setlength{\tabcolsep}{3pt}{
       \begin{tabular}{l|cccccccc|c}
        \toprule
         & {MNLI} & {QQP}  & {QNLI} & {SST-2} & {STS-B} & {MRPC} & {SQuAD$_{1.1}$} & {SQuAD$_{2.0}$} & {Avg. Diff} \\ 
        \midrule         
        Baseline & 84.53 &	91.00 &	91.41 &	93.57 &	88.90  & 86.27 & 88.48 &	76.82 & \\
        \midrule
        Mask Search & 81.21 &	89.99 &	88.38 &	92.13 &	87.10 & 83.14  &	82.66 &	71.12 & \\
        + Mask Rearrangement & 81.81 &	90.08 &	88.77 &	92.09 &	87.68 & 83.23  &	84.47 &	72.38 & + 0.60 \\
        \ha + Mask Tuning & \textbf{82.51} &	\textbf{90.35} &	\textbf{90.06} &	\textbf{92.49} &	\textbf{88.00} & \textbf{85.27}  &	\textbf{86.72} &	\textbf{75.26} & + 1.27 \\
        \bottomrule
        \end{tabular} 
    }
    }
\end{table*}

\vspace{-1mm}
\subsection{Discussion} 
\label{subsection:analysis}
\vspace{-1mm}

\textbf{Ablation Studies.}
\tref{tab:ablation} lists an ablation of the mask rearrangement (\sref{subsec:rearrange}) and tuning (\sref{subsec:tuning}) stages for pruned BERT$_\textsc{base}$ with 60\% of FLOPs.
We find that both stages help recover the baseline accuracy, and that mask tuning is in particular critical, recovering up to 2.88\% accuracy.

To further investigate the importance of the mask search and rearrangement, we compare the \textit{retraining-free} performance of the binary masks obtained by our method and other pruning criteria: weight magnitude and the gradient-based method used in DynaBERT.
We uniformly pruned the layers using the two criteria with different width multipliers.
\fref{fig:effectiveness} shows that the two methods significantly degrade the accuracy under the low sparsity regimes.
Even with mask tuning, the accuracy is not fully recovered.
The results demonstrate that our mask search and re-arrangement are necessary to get optimal binary masks, and that mask tuning is only effective when the binary mask preserves high accuracy.
More ablation studies can be found in \sref{appendix:importance_metric} and \sref{appendix:abl-search-rearrange}.

\textbf{Time Breakdown.}
We break down our pruning pipeline into 4 parts---gradient computation, mask search, rearrangement, and tuning---and we measure the latency for each stage as \tref{tab:time_breakdown} (\sref{appendix:time_breakdown}).
For GLUE and SQuAD tasks, our framework finishes in 39 and 135 seconds on average.

\vspace{-5mm}
\section{Conclusion}
\label{sec:conclusion}
\vspace{-3mm}

In this work, we have proposed a novel post-training pruning framework for Transformers that does not require model retraining.
The core of our framework is the three-stage decomposition of the pruning process.
It uses a fast Fisher-based mask search algorithm to decide which heads/filters to prune, rearranges the pruned heads/filters, and tunes the mask variables to recover the output signal for each layer.
We empirically evaluate our framework using BERT$_\textsc{base}$ and DistilBERT, where our pruning method achieves up to 50\% FLOPs reduction within only 1\% accuracy degradation on GLUE and SQuAD datasets.
This results in up to 1.56$\times$ latency speedup on an NVIDIA V100 GPU.
Importantly, our end-to-end pruning pipeline only needs 39 and 135 seconds for GLUE and SQuAD, which is 2$-$3 orders of magnitude faster than the prior methods.
\vspace{-3mm}

\begin{ack}
\vspace{-2mm}
The authors would like to thank Suhong Moon who helped with brainstorming.
We also acknowledge gracious support from 
Google Cloud, Google TRC team, and specifically Jonathan Caton, Prof. David Patterson,
and Jing Li.
Prof. Keutzer's lab is sponsored by Samsung, Intel corporation, Intel VLAB team, Intel One-API center of
excellence, as well as funding through BDD and BAIR.
Woosuk Kwon and Sehoon Kim acknowledge the support from Korea Foundation for Advanced Studies.
Amir Gholami was supported through funding from Samsung SAIT.
Michael W. Mahoney would also like to acknowledge the UC Berkeley CLTC, ARO, NSF, and ONR.
Our conclusions do not necessarily reflect the position or the policy of our sponsors, and no official endorsement should be~inferred.
\end{ack}

{
\small
\bibliographystyle{plain}
\bibliography{reference.bib}

\begin{thebibliography}{10}

\bibitem{aflalo2020knapsack}
Yonathan Aflalo, Asaf Noy, Ming Lin, Itamar Friedman, and Lihi Zelnik.
\newblock Knapsack pruning with inner distillation.
\newblock {\em arXiv preprint arXiv:2002.08258}, 2020.

\bibitem{baevski2020wav2vec}
Alexei Baevski, Henry Zhou, Abdelrahman Mohamed, and Michael Auli.
\newblock wav2vec 2.0: A framework for self-supervised learning of speech
  representations.
\newblock {\em arXiv preprint arXiv:2006.11477}, 2020.

\bibitem{banner2018post}
Ron Banner, Yury Nahshan, Elad Hoffer, and Daniel Soudry.
\newblock Post-training 4-bit quantization of convolution networks for
  rapid-deployment.
\newblock {\em arXiv preprint arXiv:1810.05723}, 2018.

\bibitem{brown2020language}
Tom~B Brown, Benjamin Mann, Nick Ryder, Melanie Subbiah, Jared Kaplan, Prafulla
  Dhariwal, Arvind Neelakantan, Pranav Shyam, Girish Sastry, Amanda Askell,
  et~al.
\newblock Language models are few-shot learners.
\newblock {\em arXiv preprint arXiv:2005.14165}, 2020.

\bibitem{cer2017semeval}
Daniel Cer, Mona Diab, Eneko Agirre, Inigo Lopez-Gazpio, and Lucia Specia.
\newblock Semeval-2017 task 1: Semantic textual similarity-multilingual and
  cross-lingual focused evaluation.
\newblock {\em arXiv preprint arXiv:1708.00055}, 2017.

\bibitem{chen2020adabert}
Daoyuan Chen, Yaliang Li, Minghui Qiu, Zhen Wang, Bofang Li, Bolin Ding, Hongbo
  Deng, Jun Huang, Wei Lin, and Jingren Zhou.
\newblock Adabert: Task-adaptive bert compression with differentiable neural
  architecture search.
\newblock {\em arXiv preprint arXiv:2001.04246}, 2020.

\bibitem{chen2021wavlm}
Sanyuan Chen, Chengyi Wang, Zhengyang Chen, Yu~Wu, Shujie Liu, Zhuo Chen, Jinyu
  Li, Naoyuki Kanda, Takuya Yoshioka, Xiong Xiao, et~al.
\newblock Wavlm: Large-scale self-supervised pre-training for full stack speech
  processing.
\newblock {\em arXiv preprint arXiv:2110.13900}, 2021.

\bibitem{chen2021chasing}
Tianlong Chen, Yu~Cheng, Zhe Gan, Lu~Yuan, Lei Zhang, and Zhangyang Wang.
\newblock Chasing sparsity in vision transformers: An end-to-end exploration.
\newblock {\em Advances in Neural Information Processing Systems},
  34:19974--19988, 2021.

\bibitem{chen2020lottery}
Tianlong Chen, Jonathan Frankle, Shiyu Chang, Sijia Liu, Yang Zhang, Zhangyang
  Wang, and Michael Carbin.
\newblock The lottery ticket hypothesis for pre-trained {BERT} networks.
\newblock {\em arXiv preprint arXiv:2007.12223}, 2020.

\bibitem{chen2020earlybert}
Xiaohan Chen, Yu~Cheng, Shuohang Wang, Zhe Gan, Zhangyang Wang, and Jingjing
  Liu.
\newblock Earlybert: Efficient bert training via early-bird lottery tickets.
\newblock {\em arXiv preprint arXiv:2101.00063}, 2020.

\bibitem{dagan2005pascal}
Ido Dagan, Oren Glickman, and Bernardo Magnini.
\newblock The pascal recognising textual entailment challenge.
\newblock In {\em Machine Learning Challenges Workshop}, pages 177--190.
  Springer, 2005.

\bibitem{devlin2018bert}
Jacob Devlin, Ming-Wei Chang, Kenton Lee, and Kristina Toutanova.
\newblock Bert: Pre-training of deep bidirectional transformers for language
  understanding.
\newblock {\em arXiv preprint arXiv:1810.04805}, 2018.

\bibitem{dolan2005automatically}
William~B Dolan and Chris Brockett.
\newblock Automatically constructing a corpus of sentential paraphrases.
\newblock In {\em Proceedings of the Third International Workshop on
  Paraphrasing (IWP2005)}, 2005.

\bibitem{dosovitskiy2020image}
Alexey Dosovitskiy, Lucas Beyer, Alexander Kolesnikov, Dirk Weissenborn,
  Xiaohua Zhai, Thomas Unterthiner, Mostafa Dehghani, Matthias Minderer, Georg
  Heigold, Sylvain Gelly, et~al.
\newblock An image is worth 16x16 words: Transformers for image recognition at
  scale.
\newblock {\em arXiv preprint arXiv:2010.11929}, 2020.

\bibitem{fan2019reducing}
Angela Fan, Edouard Grave, and Armand Joulin.
\newblock Reducing transformer depth on demand with structured dropout.
\newblock {\em arXiv preprint arXiv:1909.11556}, 2019.

\bibitem{frankle2018lottery}
Jonathan Frankle and Michael Carbin.
\newblock The lottery ticket hypothesis: Finding sparse, trainable neural
  networks.
\newblock {\em arXiv preprint arXiv:1803.03635}, 2018.

\bibitem{frantar2022spdy}
Elias Frantar and Dan Alistarh.
\newblock Spdy: Accurate pruning with speedup guarantees.
\newblock {\em arXiv preprint arXiv:2201.13096}, 2022.

\bibitem{gale2019state}
Trevor Gale, Erich Elsen, and Sara Hooker.
\newblock The state of sparsity in deep neural networks.
\newblock {\em arXiv preprint arXiv:1902.09574}, 2019.

\bibitem{tflite}
{Google}.
\newblock {Tensorflow Lite:} https://www.tensorflow.org/lite, 2017.

\bibitem{guo2020accelerating}
Cong Guo, Bo~Yang Hsueh, Jingwen Leng, Yuxian Qiu, Yue Guan, Zehuan Wang,
  Xiaoying Jia, Xipeng Li, Minyi Guo, and Yuhao Zhu.
\newblock Accelerating sparse dnn models without hardware-support via tile-wise
  sparsity.
\newblock In {\em SC20: International Conference for High Performance
  Computing, Networking, Storage and Analysis}, pages 1--15. IEEE, 2020.

\bibitem{ham20203}
Tae~Jun Ham, Sung~Jun Jung, Seonghak Kim, Young~H Oh, Yeonhong Park, Yoonho
  Song, Jung-Hun Park, Sanghee Lee, Kyoung Park, Jae~W Lee, et~al.
\newblock A\^{} 3: Accelerating attention mechanisms in neural networks with
  approximation.
\newblock In {\em 2020 IEEE International Symposium on High Performance
  Computer Architecture (HPCA)}, pages 328--341. IEEE, 2020.

\bibitem{ham2021elsa}
Tae~Jun Ham, Yejin Lee, Seong~Hoon Seo, Soosung Kim, Hyunji Choi, Sung~Jun
  Jung, and Jae~W Lee.
\newblock Elsa: Hardware-software co-design for efficient, lightweight
  self-attention mechanism in neural networks.
\newblock In {\em 2021 ACM/IEEE 48th Annual International Symposium on Computer
  Architecture (ISCA)}, pages 692--705. IEEE, 2021.

\bibitem{he2017channel}
Yihui He, Xiangyu Zhang, and Jian Sun.
\newblock Channel pruning for accelerating very deep neural networks.
\newblock In {\em Proceedings of the IEEE international conference on computer
  vision}, pages 1389--1397, 2017.

\bibitem{hendrycks2016gaussian}
Dan Hendrycks and Kevin Gimpel.
\newblock Gaussian error linear units (gelus).
\newblock {\em arXiv preprint arXiv:1606.08415}, 2016.

\bibitem{hou2020dynabert}
Lu~Hou, Zhiqi Huang, Lifeng Shang, Xin Jiang, Xiao Chen, and Qun Liu.
\newblock Dynabert: Dynamic bert with adaptive width and depth.
\newblock {\em arXiv preprint arXiv:2004.04037}, 2020.

\bibitem{hsu2021hubert}
Wei-Ning Hsu, Benjamin Bolte, Yao-Hung~Hubert Tsai, Kushal Lakhotia, Ruslan
  Salakhutdinov, and Abdelrahman Mohamed.
\newblock Hubert: Self-supervised speech representation learning by masked
  prediction of hidden units.
\newblock {\em arXiv preprint arXiv:2106.07447}, 2021.

\bibitem{hubara2020improving}
Itay Hubara, Yury Nahshan, Yair Hanani, Ron Banner, and Daniel Soudry.
\newblock Improving post training neural quantization: Layer-wise calibration
  and integer programming.
\newblock {\em arXiv preprint arXiv:2006.10518}, 2020.

\bibitem{iandola2020squeezebert}
Forrest~N Iandola, Albert~E Shaw, Ravi Krishna, and Kurt~W Keutzer.
\newblock Squeezebert: What can computer vision teach nlp about efficient
  neural networks?
\newblock {\em arXiv preprint arXiv:2006.11316}, 2020.

\bibitem{openvino}
{Intel}.
\newblock {OpenVINO:} https://docs.openvino.ai/latest/index.html, 2021.

\bibitem{iyer2017first}
Shankar Iyer, Nikhil Dandekar, and Kornl Csernai.
\newblock First quora dataset release: Question pairs.(2017).
\newblock {\em URL https://data. quora.
  com/First-Quora-Dataset-Release-Question-Pairs}, 2017.

\bibitem{jiao2019tinybert}
Xiaoqi Jiao, Yichun Yin, Lifeng Shang, Xin Jiang, Xiao Chen, Linlin Li, Fang
  Wang, and Qun Liu.
\newblock Tinybert: Distilling bert for natural language understanding.
\newblock {\em arXiv preprint arXiv:1909.10351}, 2019.

\bibitem{khetan2020schubert}
Ashish Khetan and Zohar Karnin.
\newblock schubert: Optimizing elements of bert.
\newblock {\em arXiv preprint arXiv:2005.06628}, 2020.

\bibitem{kim2021bert}
Sehoon Kim, Amir Gholami, Zhewei Yao, Michael~W Mahoney, and Kurt Keutzer.
\newblock I-bert: Integer-only bert quantization.
\newblock {\em arXiv preprint arXiv:2101.01321}, 2021.

\bibitem{kim2020neuron}
Woojeong Kim, Suhyun Kim, Mincheol Park, and Geonseok Jeon.
\newblock Neuron merging: Compensating for pruned neurons.
\newblock {\em arXiv preprint arXiv:2010.13160}, 2020.

\bibitem{kitaev2019reformer}
Nikita Kitaev, Lukasz Kaiser, and Anselm Levskaya.
\newblock Reformer: The efficient transformer.
\newblock In {\em International Conference on Learning Representations}, 2019.

\bibitem{kurtic2022optimal}
Eldar Kurtic, Daniel Campos, Tuan Nguyen, Elias Frantar, Mark Kurtz, Benjamin
  Fineran, Michael Goin, and Dan Alistarh.
\newblock The optimal bert surgeon: Scalable and accurate second-order pruning
  for large language models.
\newblock {\em arXiv preprint arXiv:2203.07259}, 2022.

\bibitem{kwon2020nimble}
Woosuk Kwon, Gyeong-In Yu, Eunji Jeong, and Byung-Gon Chun.
\newblock Nimble: Lightweight and parallel gpu task scheduling for deep
  learning.
\newblock {\em arXiv preprint arXiv:2012.02732}, 2020.

\bibitem{lagunas2021block}
Fran{\c{c}}ois Lagunas, Ella Charlaix, Victor Sanh, and Alexander~M Rush.
\newblock Block pruning for faster transformers.
\newblock {\em arXiv preprint arXiv:2109.04838}, 2021.

\bibitem{lan2019albert}
Zhenzhong Lan, Mingda Chen, Sebastian Goodman, Kevin Gimpel, Piyush Sharma, and
  Radu Soricut.
\newblock Albert: A lite bert for self-supervised learning of language
  representations.
\newblock {\em arXiv preprint arXiv:1909.11942}, 2019.

\bibitem{lazarevich2021post}
Ivan Lazarevich, Alexander Kozlov, and Nikita Malinin.
\newblock Post-training deep neural network pruning via layer-wise calibration.
\newblock In {\em Proceedings of the IEEE/CVF International Conference on
  Computer Vision}, pages 798--805, 2021.

\bibitem{lecun1990optimal}
Yann LeCun, John~S Denker, and Sara~A Solla.
\newblock Optimal brain damage.
\newblock In {\em Advances in neural information processing systems}, pages
  598--605, 1990.

\bibitem{levesque2012winograd}
Hector Levesque, Ernest Davis, and Leora Morgenstern.
\newblock The winograd schema challenge.
\newblock In {\em Thirteenth International Conference on the Principles of
  Knowledge Representation and Reasoning}. Citeseer, 2012.

\bibitem{li2020efficient}
Bingbing Li, Zhenglun Kong, Tianyun Zhang, Ji~Li, Zhengang Li, Hang Liu, and
  Caiwen Ding.
\newblock Efficient transformer-based large scale language representations
  using hardware-friendly block structured pruning.
\newblock {\em arXiv preprint arXiv:2009.08065}, 2020.

\bibitem{lin2020pruning}
Zi~Lin, Jeremiah~Zhe Liu, Zi~Yang, Nan Hua, and Dan Roth.
\newblock Pruning redundant mappings in transformer models via
  spectral-normalized identity prior.
\newblock {\em arXiv preprint arXiv:2010.01791}, 2020.

\bibitem{liu2021group}
Liyang Liu, Shilong Zhang, Zhanghui Kuang, Aojun Zhou, Jing-Hao Xue, Xinjiang
  Wang, Yimin Chen, Wenming Yang, Qingmin Liao, and Wayne Zhang.
\newblock Group fisher pruning for practical network compression.
\newblock In {\em International Conference on Machine Learning}, pages
  7021--7032. PMLR, 2021.

\bibitem{liu2019roberta}
Yinhan Liu, Myle Ott, Naman Goyal, Jingfei Du, Mandar Joshi, Danqi Chen, Omer
  Levy, Mike Lewis, Luke Zettlemoyer, and Veselin Stoyanov.
\newblock Roberta: A robustly optimized bert pretraining approach.
\newblock {\em arXiv preprint arXiv:1907.11692}, 2019.

\bibitem{liu2021rosita}
Yuanxin Liu, Zheng Lin, and Fengcheng Yuan.
\newblock Rosita: Refined bert compression with integrated techniques.
\newblock In {\em Proceedings of the AAAI Conference on Artificial
  Intelligence}, volume~35, pages 8715--8722, 2021.

\bibitem{liu2021swin}
Ze~Liu, Yutong Lin, Yue Cao, Han Hu, Yixuan Wei, Zheng Zhang, Stephen Lin, and
  Baining Guo.
\newblock Swin transformer: Hierarchical vision transformer using shifted
  windows.
\newblock {\em arXiv preprint arXiv:2103.14030}, 2021.

\bibitem{liu2021ebert}
Zejian Liu, Fanrong Li, Gang Li, and Jian Cheng.
\newblock Ebert: Efficient bert inference with dynamic structured pruning.
\newblock In {\em Findings of the Association for Computational Linguistics:
  ACL-IJCNLP 2021}, pages 4814--4823, 2021.

\bibitem{ma2020rammer}
Lingxiao Ma, Zhiqiang Xie, Zhi Yang, Jilong Xue, Youshan Miao, Wei Cui,
  Wenxiang Hu, Fan Yang, Lintao Zhang, and Lidong Zhou.
\newblock Rammer: Enabling holistic deep learning compiler optimizations with
  rtasks.
\newblock In {\em 14th $\{$USENIX$\}$ Symposium on Operating Systems Design and
  Implementation ($\{$OSDI$\}$ 20)}, pages 881--897, 2020.

\bibitem{michel2019sixteen}
Paul Michel, Omer Levy, and Graham Neubig.
\newblock Are sixteen heads really better than one?
\newblock {\em arXiv preprint arXiv:1905.10650}, 2019.

\bibitem{molchanov2019importance}
Pavlo Molchanov, Arun Mallya, Stephen Tyree, Iuri Frosio, and Jan Kautz.
\newblock Importance estimation for neural network pruning.
\newblock In {\em Proceedings of the IEEE/CVF Conference on Computer Vision and
  Pattern Recognition}, pages 11264--11272, 2019.

\bibitem{mussay2019data}
Ben Mussay, Margarita Osadchy, Vladimir Braverman, Samson Zhou, and Dan
  Feldman.
\newblock Data-independent neural pruning via coresets.
\newblock {\em arXiv preprint arXiv:1907.04018}, 2019.

\bibitem{nagel2020up}
Markus Nagel, Rana~Ali Amjad, Mart Van~Baalen, Christos Louizos, and Tijmen
  Blankevoort.
\newblock Up or down? adaptive rounding for post-training quantization.
\newblock In {\em International Conference on Machine Learning}, pages
  7197--7206. PMLR, 2020.

\bibitem{nishino2017cupy}
ROYUD Nishino and Shohei Hido~Crissman Loomis.
\newblock Cupy: A numpy-compatible library for nvidia gpu calculations.
\newblock {\em 31st confernce on neural information processing systems}, 151,
  2017.

\bibitem{tensorrt}
{NVIDIA}.
\newblock Tensor{RT}: https://developer.nvidia.com/tensorrt, 2018.

\bibitem{paszke2019pytorch}
Adam Paszke, Sam Gross, Francisco Massa, Adam Lerer, James Bradbury, Gregory
  Chanan, Trevor Killeen, Zeming Lin, Natalia Gimelshein, Luca Antiga, et~al.
\newblock Pytorch: An imperative style, high-performance deep learning library.
\newblock {\em Advances in neural information processing systems},
  32:8026--8037, 2019.

\bibitem{prasanna2020bert}
Sai Prasanna, Anna Rogers, and Anna Rumshisky.
\newblock When {BERT} plays the lottery, all tickets are winning.
\newblock {\em arXiv preprint arXiv:2005.00561}, 2020.

\bibitem{radu2019performance}
Valentin Radu, Kuba Kaszyk, Yuan Wen, Jack Turner, Jos{\'e} Cano, Elliot~J
  Crowley, Bj{\"o}rn Franke, Amos Storkey, and Michael O'Boyle.
\newblock Performance aware convolutional neural network channel pruning for
  embedded gpus.
\newblock In {\em 2019 IEEE International Symposium on Workload
  Characterization (IISWC)}, pages 24--34. IEEE, 2019.

\bibitem{rajpurkar2018know}
Pranav Rajpurkar, Robin Jia, and Percy Liang.
\newblock Know what you don't know: Unanswerable questions for squad.
\newblock {\em arXiv preprint arXiv:1806.03822}, 2018.

\bibitem{rajpurkar2016squad}
Pranav Rajpurkar, Jian Zhang, Konstantin Lopyrev, and Percy Liang.
\newblock {SQuAD}: 100,000+ questions for machine comprehension of text.
\newblock {\em arXiv preprint arXiv:1606.05250}, 2016.

\bibitem{sajjad2020effect}
Hassan Sajjad, Fahim Dalvi, Nadir Durrani, and Preslav Nakov.
\newblock On the effect of dropping layers of pre-trained transformer models.
\newblock {\em arXiv preprint arXiv:2004.03844}, 2020.

\bibitem{sanh2019distilbert}
Victor Sanh, Lysandre Debut, Julien Chaumond, and Thomas Wolf.
\newblock Distilbert, a distilled version of bert: smaller, faster, cheaper and
  lighter.
\newblock {\em arXiv preprint arXiv:1910.01108}, 2019.

\bibitem{sanh2020movement}
Victor Sanh, Thomas Wolf, and Alexander~M Rush.
\newblock Movement pruning: Adaptive sparsity by fine-tuning.
\newblock {\em arXiv preprint arXiv:2005.07683}, 2020.

\bibitem{shen2021halp}
Maying Shen, Hongxu Yin, Pavlo Molchanov, Lei Mao, Jianna Liu, and Jose~M
  Alvarez.
\newblock Halp: Hardware-aware latency pruning.
\newblock {\em arXiv preprint arXiv:2110.10811}, 2021.

\bibitem{shen2020q}
Sheng Shen, Zhen Dong, Jiayu Ye, Linjian Ma, Zhewei Yao, Amir Gholami,
  Michael~W Mahoney, and Kurt Keutzer.
\newblock Q-bert: Hessian based ultra low precision quantization of bert.
\newblock In {\em Proceedings of the AAAI Conference on Artificial
  Intelligence}, volume~34, pages 8815--8821, 2020.

\bibitem{so2019evolved}
David So, Quoc Le, and Chen Liang.
\newblock The evolved transformer.
\newblock In {\em International Conference on Machine Learning}, pages
  5877--5886. PMLR, 2019.

\bibitem{so2021primer}
David~R So, Wojciech Ma{\'n}ke, Hanxiao Liu, Zihang Dai, Noam Shazeer, and
  Quoc~V Le.
\newblock Primer: Searching for efficient transformers for language modeling.
\newblock {\em arXiv preprint arXiv:2109.08668}, 2021.

\bibitem{socher2013recursive}
Richard Socher, Alex Perelygin, Jean Wu, Jason Chuang, Christopher~D Manning,
  Andrew~Y Ng, and Christopher Potts.
\newblock Recursive deep models for semantic compositionality over a sentiment
  treebank.
\newblock In {\em Proceedings of the 2013 conference on empirical methods in
  natural language processing}, pages 1631--1642, 2013.

\bibitem{srinivas2015data}
Suraj Srinivas and R~Venkatesh Babu.
\newblock Data-free parameter pruning for deep neural networks.
\newblock {\em arXiv preprint arXiv:1507.06149}, 2015.

\bibitem{sun2019patient}
Siqi Sun, Yu~Cheng, Zhe Gan, and Jingjing Liu.
\newblock Patient knowledge distillation for bert model compression.
\newblock {\em arXiv preprint arXiv:1908.09355}, 2019.

\bibitem{sun2020mobilebert}
Zhiqing Sun, Hongkun Yu, Xiaodan Song, Renjie Liu, Yiming Yang, and Denny Zhou.
\newblock Mobilebert: a compact task-agnostic bert for resource-limited
  devices.
\newblock {\em arXiv preprint arXiv:2004.02984}, 2020.

\bibitem{tambe2021edgebert}
Thierry Tambe, Coleman Hooper, Lillian Pentecost, Tianyu Jia, En-Yu Yang, Marco
  Donato, Victor Sanh, Paul Whatmough, Alexander~M Rush, David Brooks, et~al.
\newblock Edgebert: Sentence-level energy optimizations for latency-aware
  multi-task nlp inference.
\newblock In {\em MICRO-54: 54th Annual IEEE/ACM International Symposium on
  Microarchitecture}, pages 830--844, 2021.

\bibitem{theis2018faster}
Lucas Theis, Iryna Korshunova, Alykhan Tejani, and Ferenc Husz{\'a}r.
\newblock Faster gaze prediction with dense networks and fisher pruning.
\newblock {\em arXiv preprint arXiv:1801.05787}, 2018.

\bibitem{touvron2021training}
Hugo Touvron, Matthieu Cord, Matthijs Douze, Francisco Massa, Alexandre
  Sablayrolles, and Herv{\'e} J{\'e}gou.
\newblock Training data-efficient image transformers \& distillation through
  attention.
\newblock In {\em International Conference on Machine Learning}, pages
  10347--10357. PMLR, 2021.

\bibitem{vaswani2017attention}
Ashish Vaswani, Noam Shazeer, Niki Parmar, Jakob Uszkoreit, Llion Jones,
  Aidan~N Gomez, {\L}ukasz Kaiser, and Illia Polosukhin.
\newblock Attention is all you need.
\newblock In {\em Advances in neural information processing systems}, pages
  5998--6008, 2017.

\bibitem{voita2019analyzing}
Elena Voita, David Talbot, Fedor Moiseev, Rico Sennrich, and Ivan Titov.
\newblock Analyzing multi-head self-attention: Specialized heads do the heavy
  lifting, the rest can be pruned.
\newblock {\em arXiv preprint arXiv:1905.09418}, 2019.

\bibitem{wang2018glue}
Alex Wang, Amanpreet Singh, Julian Michael, Felix Hill, Omer Levy, and Samuel~R
  Bowman.
\newblock {GLUE}: A multi-task benchmark and analysis platform for natural
  language understanding.
\newblock {\em arXiv preprint arXiv:1804.07461}, 2018.

\bibitem{wang2020hat}
Hanrui Wang, Zhanghao Wu, Zhijian Liu, Han Cai, Ligeng Zhu, Chuang Gan, and
  Song Han.
\newblock Hat: Hardware-aware transformers for efficient natural language
  processing.
\newblock {\em arXiv preprint arXiv:2005.14187}, 2020.

\bibitem{wang2021spatten}
Hanrui Wang, Zhekai Zhang, and Song Han.
\newblock Spatten: Efficient sparse attention architecture with cascade token
  and head pruning.
\newblock In {\em 2021 IEEE International Symposium on High-Performance
  Computer Architecture (HPCA)}, pages 97--110. IEEE, 2021.

\bibitem{wang2020linformer}
Sinong Wang, Belinda Li, Madian Khabsa, Han Fang, and Hao Ma.
\newblock Linformer: Self-attention with linear complexity.
\newblock {\em arXiv preprint arXiv:2006.04768}, 2020.

\bibitem{wang2020minilm}
Wenhui Wang, Furu Wei, Li~Dong, Hangbo Bao, Nan Yang, and Ming Zhou.
\newblock Minilm: Deep self-attention distillation for task-agnostic
  compression of pre-trained transformers.
\newblock {\em arXiv preprint arXiv:2002.10957}, 2020.

\bibitem{wang2019structured}
Ziheng Wang, Jeremy Wohlwend, and Tao Lei.
\newblock Structured pruning of large language models.
\newblock {\em arXiv preprint arXiv:1910.04732}, 2019.

\bibitem{warstadt2019neural}
Alex Warstadt, Amanpreet Singh, and Samuel~R Bowman.
\newblock Neural network acceptability judgments.
\newblock {\em Transactions of the Association for Computational Linguistics},
  7:625--641, 2019.

\bibitem{williams2017broad}
Adina Williams, Nikita Nangia, and Samuel~R Bowman.
\newblock A broad-coverage challenge corpus for sentence understanding through
  inference.
\newblock {\em arXiv preprint arXiv:1704.05426}, 2017.

\bibitem{wolf2020transformers}
Thomas Wolf, Julien Chaumond, Lysandre Debut, Victor Sanh, Clement Delangue,
  Anthony Moi, Pierric Cistac, Morgan Funtowicz, Joe Davison, Sam Shleifer,
  et~al.
\newblock Transformers: State-of-the-art natural language processing.
\newblock In {\em Proceedings of the 2020 Conference on Empirical Methods in
  Natural Language Processing: System Demonstrations}, pages 38--45, 2020.

\bibitem{wu2020lite}
Zhanghao Wu, Zhijian Liu, Ji~Lin, Yujun Lin, and Song Han.
\newblock Lite transformer with long-short range attention.
\newblock {\em arXiv preprint arXiv:2004.11886}, 2020.

\bibitem{xia2022structured}
Mengzhou Xia, Zexuan Zhong, and Danqi Chen.
\newblock Structured pruning learns compact and accurate models.
\newblock In {\em Proceedings of the 60th Annual Meeting of the Association for
  Computational Linguistics (Volume 1: Long Papers)}, pages 1513--1528, 2022.

\bibitem{xin2020deebert}
Ji~Xin, Raphael Tang, Jaejun Lee, Yaoliang Yu, and Jimmy Lin.
\newblock Deebert: Dynamic early exiting for accelerating bert inference.
\newblock In {\em Proceedings of the 58th Annual Meeting of the Association for
  Computational Linguistics}, pages 2246--2251, 2020.

\bibitem{xu2021bert}
Jin Xu, Xu~Tan, Renqian Luo, Kaitao Song, Jian Li, Tao Qin, and Tie-Yan Liu.
\newblock Nas-bert: Task-agnostic and adaptive-size bert compression with
  neural architecture search.
\newblock {\em arXiv preprint arXiv:2105.14444}, 2021.

\bibitem{yang2019xlnet}
Zhilin Yang, Zihang Dai, Yiming Yang, Jaime Carbonell, Russ~R Salakhutdinov,
  and Quoc~V Le.
\newblock Xlnet: Generalized autoregressive pretraining for language
  understanding.
\newblock {\em Advances in neural information processing systems}, 32, 2019.

\bibitem{yao2021mlpruning}
Zhewei Yao, Linjian Ma, Sheng Shen, Kurt Keutzer, and Michael~W Mahoney.
\newblock Mlpruning: A multilevel structured pruning framework for
  transformer-based models.
\newblock {\em arXiv preprint arXiv:2105.14636}, 2021.

\bibitem{yin2021autotinybert}
Yichun Yin, Cheng Chen, Lifeng Shang, Xin Jiang, Xiao Chen, and Qun Liu.
\newblock Autotinybert: Automatic hyper-parameter optimization for efficient
  pre-trained language models.
\newblock {\em arXiv preprint arXiv:2107.13686}, 2021.

\bibitem{yvinec2021red}
Edouard Yvinec, Arnaud Dapogny, Matthieu Cord, and Kevin Bailly.
\newblock Red: Looking for redundancies for data-freestructured compression of
  deep neural networks.
\newblock {\em Advances in Neural Information Processing Systems},
  34:20863--20873, 2021.

\bibitem{zadeh2020gobo}
Ali~Hadi Zadeh, Isak Edo, Omar~Mohamed Awad, and Andreas Moshovos.
\newblock Gobo: Quantizing attention-based nlp models for low latency and
  energy efficient inference.
\newblock In {\em 2020 53rd Annual IEEE/ACM International Symposium on
  Microarchitecture (MICRO)}, pages 811--824. IEEE, 2020.

\bibitem{zafrir2019q8bert}
Ofir Zafrir, Guy Boudoukh, Peter Izsak, and Moshe Wasserblat.
\newblock Q8bert: Quantized 8bit bert.
\newblock {\em arXiv preprint arXiv:1910.06188}, 2019.

\bibitem{zhao2019improving}
Ritchie Zhao, Yuwei Hu, Jordan Dotzel, Christopher De~Sa, and Zhiru Zhang.
\newblock Improving neural network quantization without retraining using
  outlier channel splitting.
\newblock {\em Proceedings of Machine Learning Research}, 2019.

\bibitem{zhou2020bert}
Wangchunshu Zhou, Canwen Xu, Tao Ge, Julian McAuley, Ke~Xu, and Furu Wei.
\newblock Bert loses patience: Fast and robust inference with early exit.
\newblock {\em Advances in Neural Information Processing Systems},
  33:18330--18341, 2020.

\end{thebibliography}
}

\clearpage

\appendix

\section{Appendix}

\subsection{\textbf{Proof of Equation~\ref{eqn:from_theorem:optimal_mask}}}
\label{appendix:optimality_proof}

We prove 
\eref{eqn:from_theorem:optimal_mask}
by contradiction.
Suppose that there exists a mask $\text{m}$ such that 
\begin{equation}
\small
    \sum_{i \in Z(\text{m})} \mathcal{I}_{ii} < \sum_{i \in Z(\text{m}^\ast)} \mathcal{I}_{ii} \label{eq:alg1_pf},
\end{equation}
where the mask $\text{m}^\ast$ is the output of Algorithm~\ref{alg:search_flops}.
Let $h = || \text{m}^{\textsc{mha}} ||_0$, i.e. the total number of heads in the architecture pruned by the mask $\text{m}$.
Then we construct a new mask $\text{m}'$ as follows:
\begin{enumerate}
    \item Keep the mask for MHA layers.
    That is, $\text{m}'^{\textsc{mha}} = \text{m}^{\textsc{mha}}$.
    \item Construct $\text{m}'^{\textsc{ffn}}$ as in the inner statements of the for loop (line~\ref{alg:search_flops:for} of Algorithm~\ref{alg:search_flops}).
    That is, given a mask initialized to $\mathbb{1}$, we zero out the $k$ filter mask variables with the least important scores, where $k = LN - \lfloor (C - f \text{F}_{\text{head}}) / \text{F}_\text{filter} \rfloor$.
\end{enumerate}
Obviously, the mask $\text{m}'$ satisfies the FLOPs constraint~\eref{eq:flops_objective}.
Moreover, as the mask $\text{m}'$ prunes the least important filters, the following two inequalities hold:
\begin{gather}
    \sum_{i \in Z(\text{m}'^{\textsc{ffn}})} \mathcal{I}_{ii} \leq \sum_{i \in Z(\text{m}^{\textsc{ffn}})} \mathcal{I}_{ii}, \\
    \sum_{i \in Z(\text{m}')} \mathcal{I}_{ii} = \sum_{i \in Z(\text{m}'^{\textsc{mha}})} \mathcal{I}_{ii} + \sum_{i \in Z(\text{m}'^{\textsc{ffn}})} \mathcal{I}_{ii} \leq \sum_{i \in Z(\text{m}^{\textsc{mha}})} \mathcal{I}_{ii} + \sum_{i \in Z(\text{m}^{\textsc{ffn}})} \mathcal{I}_{ii} = \sum_{i \in Z(\text{m})} \mathcal{I}_{ii}.
\end{gather}
Then we construct another mask $\text{m}^\star$ from $\text{m}'$ such that:
\begin{enumerate}
    \item Keep the mask for FFN layers.
    That is, $\text{m}^{\star \textsc{ffn}} = \text{m}'^{\textsc{ffn}}$.
    \item Construct $\text{m}^{\star \textsc{mha}}$ by zeroing out $h$ head mask variables with the least important scores from a mask initialized to $\mathbb{1}$.
\end{enumerate}
Due to the observation (2) in~\sref{subsec:search}, the following inequalities hold:
\begin{gather}
    \sum_{i \in Z(\text{m}^{\star \textsc{mha}})} \mathcal{I}_{ii} \leq \sum_{i \in Z(\text{m}'^{\textsc{mha}})} \mathcal{I}_{ii}, \\
    \sum_{i \in Z(\text{m}^\star)} \mathcal{I}_{ii} = \sum_{i \in Z(\text{m}^{\star \textsc{mha}})} \mathcal{I}_{ii} + \sum_{i \in Z(\text{m}^{\star \textsc{ffn}})} \mathcal{I}_{ii} \leq \sum_{i \in Z(\text{m}'^{\textsc{mha}})} \mathcal{I}_{ii} + \sum_{i \in Z(\text{m}'^{\textsc{ffn}})} \mathcal{I}_{ii} = \sum_{i \in Z(\text{m}')} \mathcal{I}_{ii}.
\end{gather}
Essentially, $\text{m}^\star$ is the mask when $n$ (in line~\ref{alg:search_flops:for} of Algorithm~\ref{alg:search_flops}) is $h$.
As Algorithm~\ref{alg:search_flops} finds the minimum by iterating different values of $n$, the following inequalities hold:
\begin{gather}
    \sum_{i \in Z(\text{m}^\ast)} \mathcal{I}_{ii} \leq \sum_{i \in Z(\text{m}^\star)} \mathcal{I}_{ii}. 
\end{gather}
Finally, the above inequalities are combined as follows:
\begin{gather}
    \sum_{i \in Z(\text{m}^\ast)} \mathcal{I}_{ii} \leq \sum_{i \in Z(\text{m}^\star)} \mathcal{I}_{ii} \leq \sum_{i \in Z(\text{m}')} \mathcal{I}_{ii} \leq \sum_{i \in Z(\text{m})} \mathcal{I}_{ii},
\end{gather}
which contradicts~\eref{eq:alg1_pf}.
Thus, the output mask $\text{m}^\ast$ of Algorithm~\ref{alg:search_flops} is the minimizer of~\eref{eq:flops_objective}.
\hfill $\square$

\subsection{\textbf{Latency-aware Search Algorithm}}
\label{appendix:latency_algorithm}

Algorithm~\ref{alg:search_latency} is our proposed algorithm for latency-aware mask search, which extends Algorithm~\ref{alg:search_flops}.
It takes as inputs the given latency constraint, approximated $\text{LAT}$ functions for MHA and FFN layers, and the diagonal Fisher information matrix $\mathcal{I}$.
Overall, Algorithm~\ref{alg:search_latency} has the same structure as Algorithm~\ref{alg:search_flops}.
A notable difference between the two is that Algorithm~\ref{alg:search_latency} separately considers the constant part (i.e., when the number of heads/filters is below the threshold $T$) in line 1--5.
Another difference is that Algorithm~\ref{alg:search_latency} uses $a_\text{head}$ and $a_\text{filter}$ instead of $\text{F}_\text{head}$ and $\text{F}_\text{filter}$ in Algorithm~\ref{alg:search_flops}.

\begin{algorithm}[tb]
   \caption{Mask Search with a Latency Constraint}
   \label{alg:search_latency}
   
\footnotesize
    {\bfseries Input:} Latency constraint $C$, LAT function parameters ($a_{\text{head}}$,$c_{\text{head}}$,$T_{\text{head}}$),($a_{\text{filter}}$,$c_{\text{filter}}$,$T_{\text{filter}}$), diagonal Fisher information matrix $\mathcal{I}$

\begin{algorithmic}[1]
    \STATE $\text{HI} = $ indicies of $T_\text{head}$ most important heads in each layer
    \STATE $\text{FI} = $ indicies of $T_\text{filter}$ most important filters in each layer

    \STATE $H' = H - T_\text{head}$
    \STATE $N' = N - T_\text{filter}$
    \STATE $C' = C - L (c_\text{head} + c_\text{filter})$

    \STATE \textbf{for} {$n=0$ {\bfseries to} $LH'$} \textbf{do}
        \STATE \quad $I = $ indicies of $n$ most important heads not in $\text{HI}$
        \STATE \quad $f = \lfloor (C' - n a_{\text{head}}) / a_\text{filter} \rfloor $
        \STATE \quad $J = $ indicies of $f$ most important filters not in $\text{FI}$
        \STATE \quad $\text{HI}' = \text{HI} \cup I$
        \STATE \quad $\text{FI}' = \text{FI} \cup J$
        \STATE \quad $S[n] = \sum_{i \in \text{HI}' \cup \text{FI}'} \mathcal{I}_{ii}$
        \STATE \quad $R[n] = (\text{HI}', \text{FI}')$
    \STATE \textbf{end for}
    \STATE $n^\ast = \argmax_n S[n]$
    \STATE $\text{HI}^\ast, \text{FI}^\ast = R[n^\ast]$
    \STATE Initialize $\text{m}^{\textsc{MHA}}$ and $\text{m}^{\textsc{FFN}}$ as 0
    \STATE $\text{m}^{\textsc{MHA}}[\text{HI}^\ast] = 1$
    \STATE $\text{m}^{\textsc{FFN}}[\text{FI}^\ast] = 1$ 
\end{algorithmic}

     {\bfseries Output:} $\text{m}^\ast = (\text{m}^{\textrm{MHA}}$, $\text{m}^{\textrm{FFN}})$

\end{algorithm}

\subsection{\textbf{Derivation of Equation~\ref{eq:layerwise_opt}}}
\label{appendix:derivation_block_diagonal}

Based on~\eref{eq:fim} and the warm-start constraint in~\sref{subsec:rearrange}, the optimization problem~\eref{eq:argmin_hessian} is written as follows:
\begin{gather}
\small
    \argmin_{\text{m}} (\mathbb{1} - \text{m})^\intercal \mathcal{I} (\mathbb{1} - \text{m}), \label{eq:argmin_fisher} \\
    \text{s.t. } || \text{m}_l ||_0 = || \text{m}^\ast_l ||_0 \ \text{for } l = 1, 2, \dots, L \label{eq:warmup_constraint},
\end{gather}
where $\text{m}^\ast$ is the mask searched in~\sref{subsec:search} using the diagonal approximation of $\mathcal{I}$.
Under the block diagonal assumption, \eref{eq:argmin_fisher} can be reduced as follows:
\begin{gather}
\small
    \argmin_{\text{m}} (\mathbb{1} - \text{m})^\intercal \mathcal{I} (\mathbb{1} - \text{m}) \approx \argmin_{\text{m}} \sum_{l=1}^L (\mathbb{1} - \text{m}_l)^\intercal \mathcal{I}_l (\mathbb{1} - \text{m}_l), \label{eq:argmin_block_sum}
\end{gather}
where $\mathcal{I}_l$ is the $l$-th diagonal block of $\mathcal{I}$.
Here what we want to show is that the problem~\eref{eq:argmin_block_sum} can be solved by independently solving the optimization problem~\eref{eq:layerwise_opt} for each layer.
We prove this by contradiction.
Suppose that $\hat{\text{m}} = ({\hat{\text{m}}}_1, \dots, {\hat{\text{m}}}_L)$ is the mask obtained by solving ~\eref{eq:layerwise_opt} for each layer.
If there exists a mask $\text{m}$ that strictly better optimizes~\eref{eq:argmin_block_sum} than $\hat{\text{m}}$:
\begin{gather}
\small
    \sum_{l=1}^L (\mathbb{1} - \text{m}_l)^\intercal \mathcal{I}_l (\mathbb{1} - \text{m}_l) < \sum_{l=1}^L (\mathbb{1} - {\hat{\text{m}}}_l)^\intercal \mathcal{I}_l (\mathbb{1} - {\hat{\text{m}}}_l),
\end{gather}
while also satisfying the constraint~\eref{eq:warmup_constraint}, then there must exist a layer $k$ such that
\begin{gather}
\small
    (\mathbb{1} - \text{m}_k)^\intercal \mathcal{I}_k (\mathbb{1} - \text{m}_k) < (\mathbb{1} - {\hat{\text{m}}}_k)^\intercal \mathcal{I}_k (\mathbb{1} - {\hat{\text{m}}}_k).
\end{gather}
However, this contradicts the assumption that ${\hat{\text{m}}}_k$ is the minimizer of~\eref{eq:layerwise_opt} for layer $k$.
Therefore, such a mask as $\text{m}$ cannot exist, and ${\hat{\text{m}}}_k$ is the optimal solution for~\eref{eq:argmin_block_sum}.


\subsection{\textbf{Formulating Equation~\ref{eq:reconstruction} as a Linear Least Squares Problem}}
\label{appendix:derivation_lsa}

For a MHA layer, the problem of minimizing reconstruction error can be written as follows:
\begin{gather}
\small
    \argmin_{\text{m}^{\textsc{mha}}_l} || \big(\text{x}  + \text{MHA}(\text{x}; \text{m}^{\textsc{mha}}_l)\big) - \big(\text{x}' + \text{MHA}(\text{x}'; \mathbb{1})\big) ||_2^2,  \\
    \text{s.t. } Z(\text{m}^{\textsc{mha}}_l) = Z({\hat{\text{m}}}^{\textsc{mha}}_l), \label{eq:reconstruction_constraint}
\end{gather}
where ${\hat{\text{m}}}$ is the mask obtained as the result of mask rearrangement (\sref{subsec:rearrange}) and
$Z(\text{m})$ denotes the indices of zero entries in $\text{m}$.
\eref{eq:reconstruction_constraint} is the constraint we impose in~\sref{subsec:tuning} that the zero-valued mask variables in ${\hat{\text{m}}}$ are fixed to 0 so that the tuned mask also satisfies the FLOPs/latency constraint.
Then we rewrite the problem as the following linear least squares problem:
\begin{gather}
\small
    \argmin_{\text{m}^{}_l} || \text{A} \text{m}^{}_l - \text{b}  ||_2^2, \label{eq:least_squares} \\
    \text{where } \ \text{A} := [\hat{m}^{\textsc{mha}}_{l,1} \attn_1(\text{x}), \dots, \hat{m}^{\textsc{mha}}_{l,H} \attn_H(\text{x})] \ \text{ and } \ \text{b} := \big(\text{x}' + \sum_{h=1}^{H} \attn_h(\text{x}')\big) - \text{x}.
\end{gather}
Here, the elements of $\hat{\text{m}}^\textsc{mha}_l$ are multiplied to the matrix $\text{A}$ to ensure that the output activations of the pruned heads are not used to reconstruct the original output.
Although \eref{eq:least_squares} has a closed form solution $ (\text{A}^\intercal \text{A})^{-1} \text{A}^\intercal \text{B}$, we use the numerical solver in CuPy for higher stability. 

\subsection{\textbf{Experimental Details}}
\label{appendix:eval_details}

\subsubsection{Experimental Setup}
Our framework is implemented on top of PyTorch v1.9.1~\cite{paszke2019pytorch} and HuggingFace Transformers v4.12.0~\cite{wolf2020transformers}.
For the baseline, we downloaded the pre-trained checkpoints from the HuggingFace Transformers repository, 
and we fine-tuned them on GLUE~\cite{wang2018glue} and SQuAD~\cite{rajpurkar2016squad, rajpurkar2018know} datasets with the standard training recipe.
We then use 2K examples from the training sets to prune the baseline models.
We report accuracy for GLUE tasks, except for STS-B that we report Spearman Correlation, and F1 score for SQuAD tasks on the development sets.
All experiments in this paper are conducted on an AWS p3.2xlarge instance which has an NVIDIA V100 GPU.
We used seed numbers from 0 to 9, and we reported the averaged results.

\subsubsection{Datasets}
GLUE tasks~\cite{wang2018glue} include sentence similarity (QQP~\cite{iyer2017first}, MRPC~\cite{dolan2005automatically}, STS-B~\cite{cer2017semeval}), 
sentiment classification (SST-2~\cite{socher2013recursive}), textual entailment (RTE~\cite{dagan2005pascal}) and natural language inference (MNLI~\cite{williams2017broad}, QNLI~\cite{rajpurkar2016squad}).
There are 364K, 4K, 6K, 67K, 3K, 392K, 105K training examples, respectively. 
We exclude CoLA~\cite{warstadt2019neural} and WLNI~\cite{levesque2012winograd} due to their unstable behaviors.

SQuAD 1.1~\cite{rajpurkar2016squad} and SQuAD 2.0~\cite{rajpurkar2018know} are question and answering tasks, each of which contains 88K and 130K training examples.
SQuAD 2.0 is an extension of SQuAD 1.1 by including unanswerable questions whose answers are not stated in the given contexts.

\subsection{Impact of Sample Dataset Size}
\label{appendix:sample_dataset}

\begin{figure*}[h]
\centering
\includegraphics[width=\textwidth]{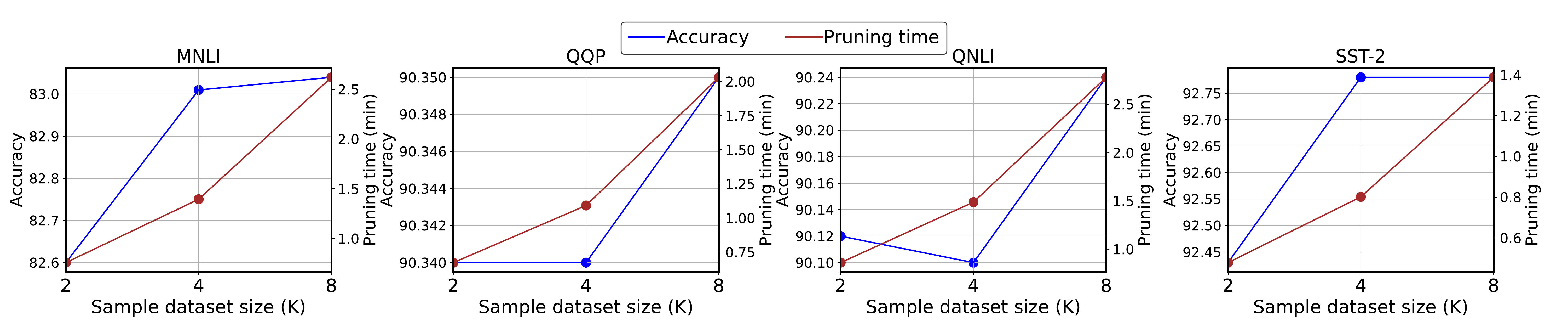}
\caption{
Accuracy and pruning time with 2K, 4K, 8K samples.
The FLOPs constraint is 60$\%$.
}
\label{fig:dataset_size}
\end{figure*}

There is a trade-off between sample dataset size and accuracy.
\fref{fig:dataset_size} shows the correlation between sample size, pruning time, and accuracy for 4 GLUE datasets (with more than 64K training examples).
Note that for simplicity we used 2K samples in all our experiments in~\sref{sec:eval}.
\fref{fig:dataset_size} demonstrates that using more examples can improve our accuracy results by up to 0.4$\%$ with 2--4$\times$ longer pruning time (which is still less than 3 minutes).

\subsection{\textbf{Details for the Comparison with the Prior Methods}}
\label{appendix:comparison_details}

\begin{table*}[h]
\caption{
The absolute accuracy and the amount of accuracy degradation from the baseline after pruning BERT$_\textsc{BASE}$ using our method and the prior structured pruning methods with different relative FLOPs.
$^{\dagger}$Reported as F1 score instead of accuracy.
}
\vskip 0.05in
\label{tab:comparison_details}
    \centering
    \small{
    \setlength{\tabcolsep}{5pt}{
       \begin{tabular}{l|c|cccc|cccc}
        \toprule
        \multirow{2}{*}{Method} & Rel. & \multicolumn{4}{c|}{Accuracy} & \multicolumn{4}{c}{Accuracy Diff}\\
         & FLOPs & QQP & QNLI & SST-2 & MRPC & QQP & QNLI & SST-2 & MRPC \\
        \midrule         
        Flop~\cite{wang2019structured} & Baseline & - & 91.6 & 92.7 & 90.9$^{\dagger}$ & - & 0 & 0 & 0 \\
        & 66.7     & - & 89.0 & 92.1 & 88.6$^{\dagger}$ & - & -2.6 & -0.6 & -2.3 \\
        \midrule
        SLIP~\cite{lin2020pruning} & Baseline & 90.6 & 91.6 & 92.7 & 90.9$^{\dagger}$ &  0 & 0 & 0 & 0 \\
        & 65.6 & 89.7 & 90.7 & 91.7 & 89.9$^{\dagger}$ & -0.9 & -0.9 & -1.0 & -1.0 \\
        & 61.5 & 88.9 & 89.5 & 91.8 & 88.1$^{\dagger}$ & -1.7 & -2.1 & -0.9 & -2.8 \\
        \midrule
        Sajjad et al.~\cite{sajjad2020effect} & Baseline & 91.1 & 91.1 & 92.4 & 88.0 & 0 & 0 & 0 & 0 \\
         & 66.7 & 90.6 & 89.7 & 90.6& 79.4 & -0.4 & -1.4	& -1.8 & -8.6 \\
         & 50.0 & 90.4 & 87.6 & 90.3 & 80.2 & -0.7 & -3.5 & -2.2 & -7.8 \\
        \midrule
        DynaBERT~\cite{hou2020dynabert} & Baseline & - & - & 92.9 & 87.7 & - & - & 0 & 0 \\
         & 75.0 & - & - & 92.3 & 86.0 & - & - & -0.6 & -1.7 \\
         & 50.0 & - & - &  91.9 & 86.0 & - & - & -1.0 & -1.7 \\ 
        \midrule
        EBERT~\cite{liu2021ebert} & Baseline & 87.9 & 91.5 & 93.2 & - & 0 & 0 & 0 & - \\
         & 60.0 & 87.5 & 90.2 & 92.2 & - & -0.4 & -1.3 & -1.0 & - \\
         & 50.0 & 87.0 & 89.6 & 91.6 & - & -0.7 & -1.9 & -1.6 & - \\
        \midrule
        BMP~\cite{lagunas2021block} & Baseline & 91.1 & - & 92.7 & - & 0 & - & 0 & - \\
         & 50.0 & 90.4 & - & 90.7 & - & -0.7 & - & -2.0 & - \\
        \midrule
        BMP~\cite{lagunas2021block} & Baseline & - & - & 92.7 & - & - & - & 0 & - \\
         (Reproduced) & 72.6 & - & - & 92.1 & - & - & - & -0.6 & - \\
         & 56.5 & - & - & 91.5 & - & - & - & -1.2 & - \\
         & 46.7 & - & - & 90.8 & - & - & - & -1.9 & - \\
        \midrule
        \ha Ours & Baseline & 91.0 & 91.4 & 93.6 & 86.3 & 0 & 0 & 0 & 0 \\
        \ha  & 70.0 & 90.7 & 90.9 & 93.0 & 86.1 & -0.3 & -0.5 & -0.6 & -0.2 \\
        \ha  & 60.0 & 90.4 & 90.0 & 92.5 & 85.3 & -0.6 & -1.4 & -1.1 & -1.0 \\
        \ha  & 50.0 & 89.5 & 88.7 & 91.6 & 83.2 & -1.5 & -2.7 & -2.0 & -3.1 \\
        \bottomrule
        \end{tabular} 
    }
    }
\end{table*}

We compare the FLOPs-accuracy trade-off of our method to the prior structured pruning methods for Transformers in Flop~\cite{wang2019structured}, SLIP~\cite{lin2020pruning}, Sajjad et al.~\cite{sajjad2020effect}, DynaBERT~\cite{hou2020dynabert}, EBERT~\cite{liu2021ebert}, and Block Movement Pruning (BMP)~\cite{lagunas2021block} on 4 GLUE tasks, QQP, QNLI, SST-2, and MRPC.
We use the FLOPs and accuracy values reported in each paper (for Flop, we use the values reported in SLIP).
To make a fair comparison, we use the experimental results \textit{without} any additional knowledge distillation or data augmentation in each paper.
Because all of these papers have slight differences in their baseline accuracy, it is difficult to directly compare the absolute accuracy for the pruned models.
Therefore, we use the amount of the accuracy drop (i.e., accuracy of the pruned model subtracted by the accuracy of the original model) instead.
For Flop and SLIP, the results for MRPC are reported as F1 score instead of accuracy. 
For these cases, we report that the amount of the F1 score drop instead, assuming that it is similar to the amount of the accuracy drop.
Since BMP only reports the parameter counts reduction, we directly use this value as FLOPs reduction, under an assumption that the head pruning ratio is similar to the filter pruning ratio.
We include the full table in~\tref{tab:comparison_details}.

\subsection{Performance at high sparsity}
\label{appendix:high_sparsity}

\begin{figure*}[h]
\centering
\includegraphics[width=0.66\textwidth]{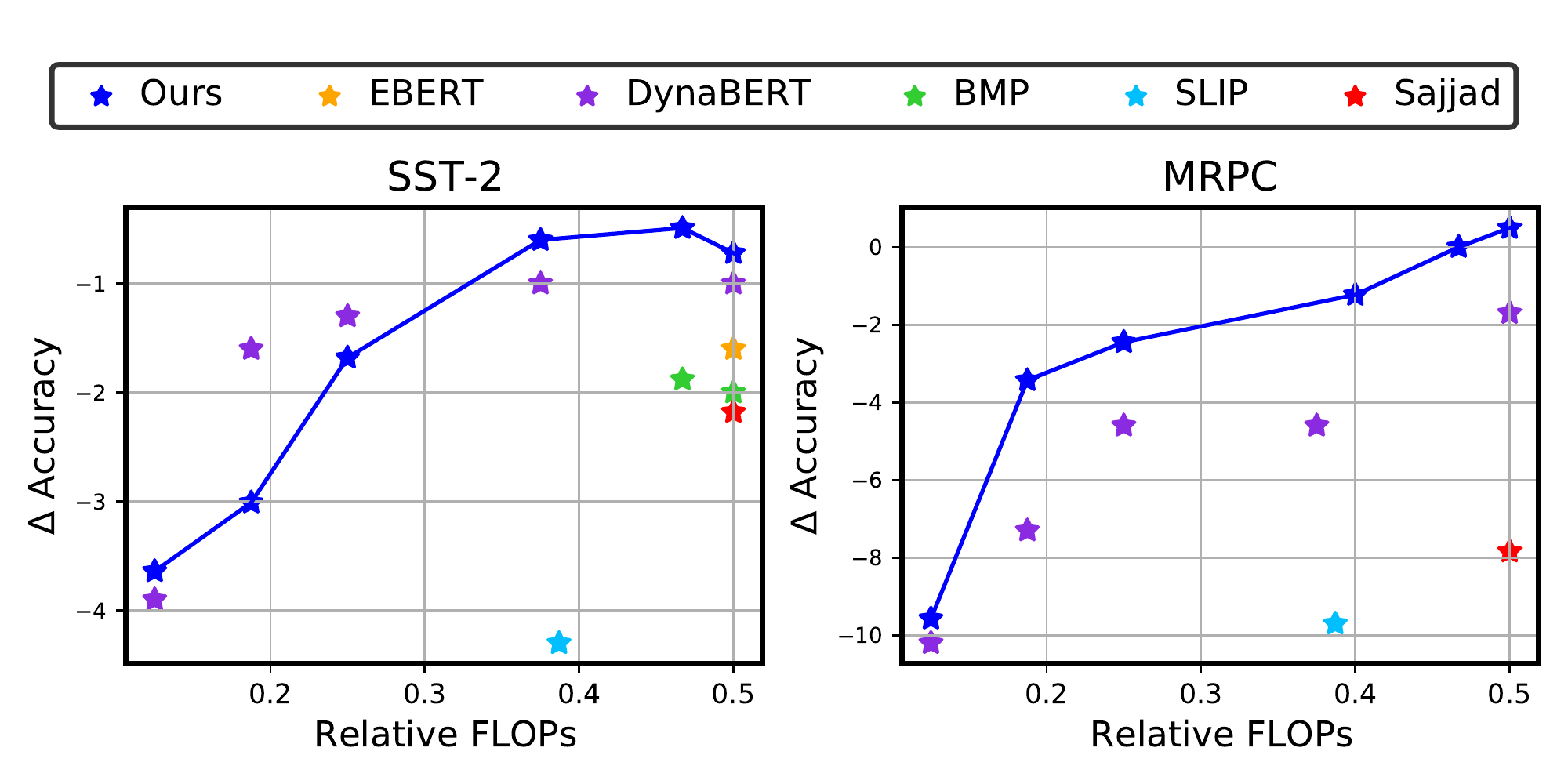}
\caption{
Performance comparison at high sparsity.
Each method (including ours) \textit{retrains} the pruned models \textit{without} data augmentation and knowledge distillation.
}
\label{fig:high_sparsity}
\end{figure*}

For high sparsity, our framework can be used with retraining to recover the accuracy.
In this experiment, our framework skips Mask Tuning (\sref{subsec:tuning}) and retrains the model parameters with the binary mask fixed.
Specifically, it retrains the pruned BERT$_\textsc{BASE}$ model for 3 epochs with the full training dataset with learning rate $2e^{-5}$.
The retraining cost can be considered the same as that of Sajjad et al. and much lower than DynaBERT~\cite{hou2020dynabert}, EBERT~\cite{liu2021ebert}, and BMP~\cite{lagunas2021block} (more details in~\sref{appendix:retraining_cost}).

\fref{fig:high_sparsity} shows the FLOPs-accuracy comparison between our method and the prior structured pruning methods at high sparsity.
For fair comparison, we used the results \textit{without} knowledge distillation and data augmentation reported in each paper.
For the SST-2 dataset, our method performs comparably to DynaBERT and outperforms all other methods.
For the MRPC dataset, our framework consistently outperforms all of the baselines.
These results imply that our Mask Search (\sref{subsec:search}) and Mask Rearrangement (\sref{subsec:rearrange}) find more optimal binary masks than other methods even at high sparsity, which is consistent with~\fref{fig:effectiveness}.

\subsection{\textbf{Retraining Costs}}
\label{appendix:retraining_cost}
DynaBERT~\cite{hou2020dynabert} consists of 2-stage training, 1 epoch of width and depth-adaptive training followed by additional 3 epochs of final fine-tuning.
For the first stage, it jointly trains 12 different width and depth configurations, roughly adding up 12$\times$ training costs than the normal training.
EBERT~\cite{liu2021ebert} consists of 2-stage training, 3 epochs that jointly trains the pruning parameters and the model weights, followed by 3 additional epochs of final fine-tuning.
BMP~\cite{lagunas2021block} requires 20 epochs of training to search for the optimal pruning configuration and retrain the model weights.
Similarly, CoFi~\cite{xia2022structured} requires 20 epochs of pruning and 20 epochs of post-pruning fine-tuning.
We measure end-to-end retraining latency on an NVIDIA V100 GPU using a batch size of 32 for all experiments.

\subsection{Pruning Time Breakdown}
\label{appendix:time_breakdown}

\begin{table}[h]
\caption{
Time breakdown (in percentage) of our pruning framework on a single NVIDIA V100 GPU.
It consists of Gradient Computation (GC), Mask Search (MS, \sref{subsec:search}), Mask Rearrangement (MR, \sref{subsec:rearrange}), and Mask Tuning (MT, \sref{subsec:tuning}).
In the last column, we provide the total amount of time for end-to-end pruning, in seconds.
}
\label{tab:time_breakdown}
    \centering
    \small{
    \setlength{\tabcolsep}{6pt}{
      \begin{tabular}{c|cccc|c}
        \toprule
        & GC & MS & MR & MT & Total (s) \\
        \midrule
        GLUE & 23.8\% & 0.3\% & 9.4\% & 66.5\% & 39.3 \\
        SQuAD & 20.5\% & 0.1\% & 3.5\% & 76.0\% & 135.1 \\
        \bottomrule
        \end{tabular} 
    }
    }
\end{table}

\subsection{Efficacy of the Fisher-based Importance Metric}
\label{appendix:importance_metric}

\begin{figure*}[h]
\centering
\includegraphics[width=0.66\textwidth]{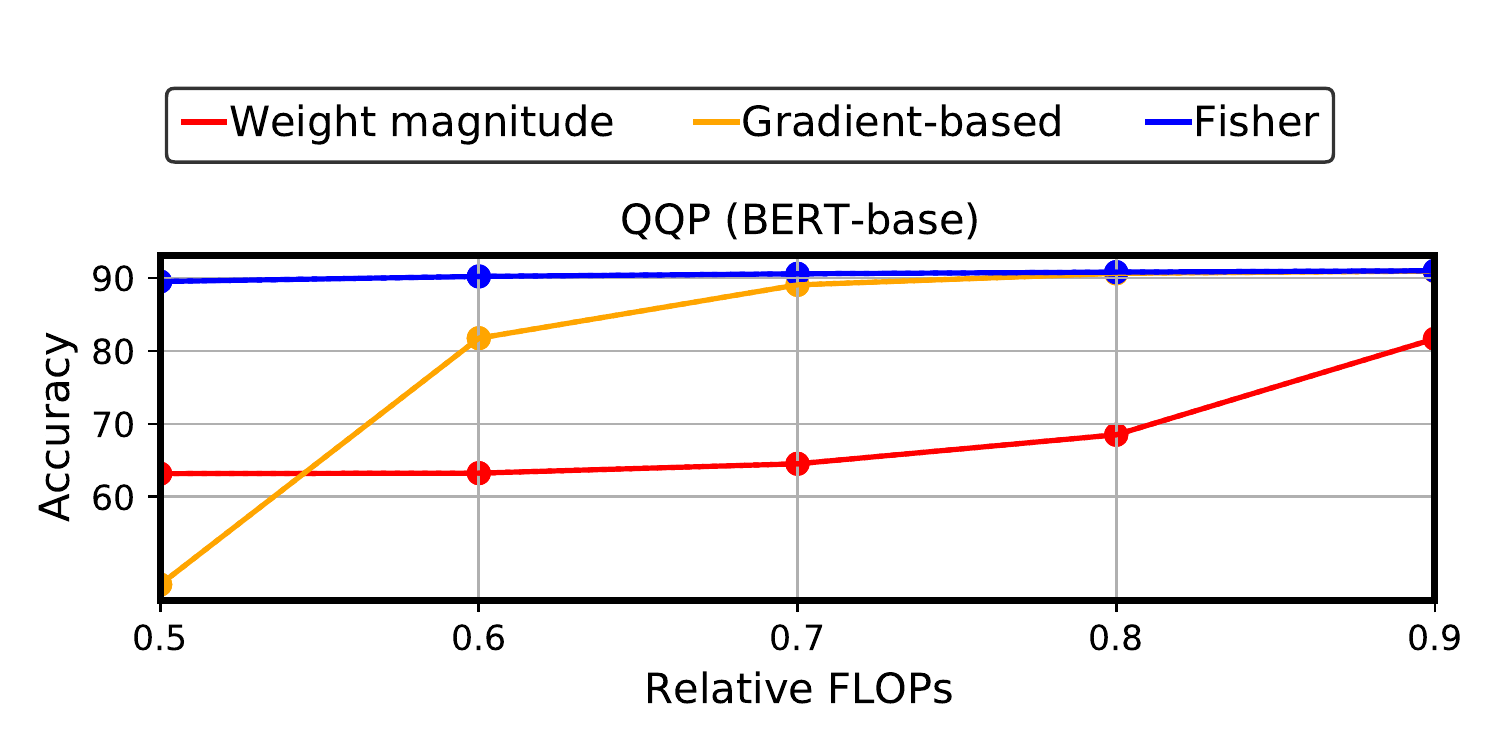}
\caption{
Pruning performance of our pipeline when different importance metrics are used. Note that the mask re-arrangement stage is \textit{skipped} in this experiment.
}
\label{fig:ablation1}
\end{figure*}

\begin{figure*}[h]
\centering
\includegraphics[width=0.66\textwidth]{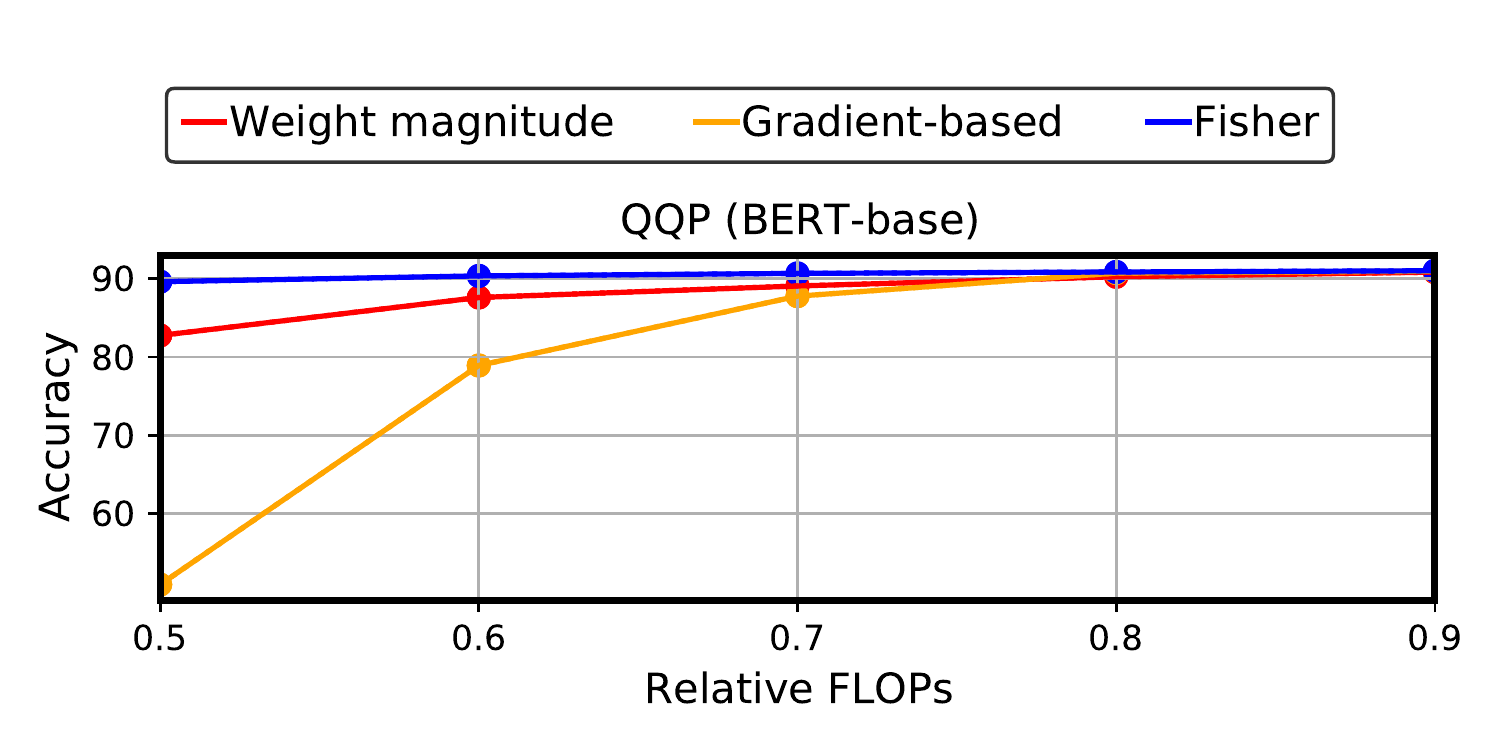}
\caption{
Pruning performance of our pipeline when different importance metrics are used. Note that the mask re-arrangement stage is \textit{included} in this experiment.
}
\label{fig:ablation2}
\end{figure*}

To demonstrate the efficacy of the Fisher-based importance score, we compare its performance with two other importance metrics (i.e., weight magnitude and gradient-based), which can be plugged into the mask search algorithm.
However, since the mask re-arrangement algorithm needs signals that capture the interaction between mask variables, the mask re-arrangement technique can only be used with the Fisher information matrix.
Hence, we designed two experiments: without and with the mask re-arrangement stage.
\fref{fig:ablation1} shows the results when the mask re-arrangement is skipped.
\fref{fig:ablation2} shows the results when the Fisher-based mask re-arrangement is applied to all.
In both experiments, our Fisher importance metric significantly outperforms the two other metrics.

\subsection{Efficacy of Mask Search and Mask Re-arrangement}
\label{appendix:abl-search-rearrange}

\begin{figure*}[h]
\centering
\includegraphics[width=0.5\textwidth]{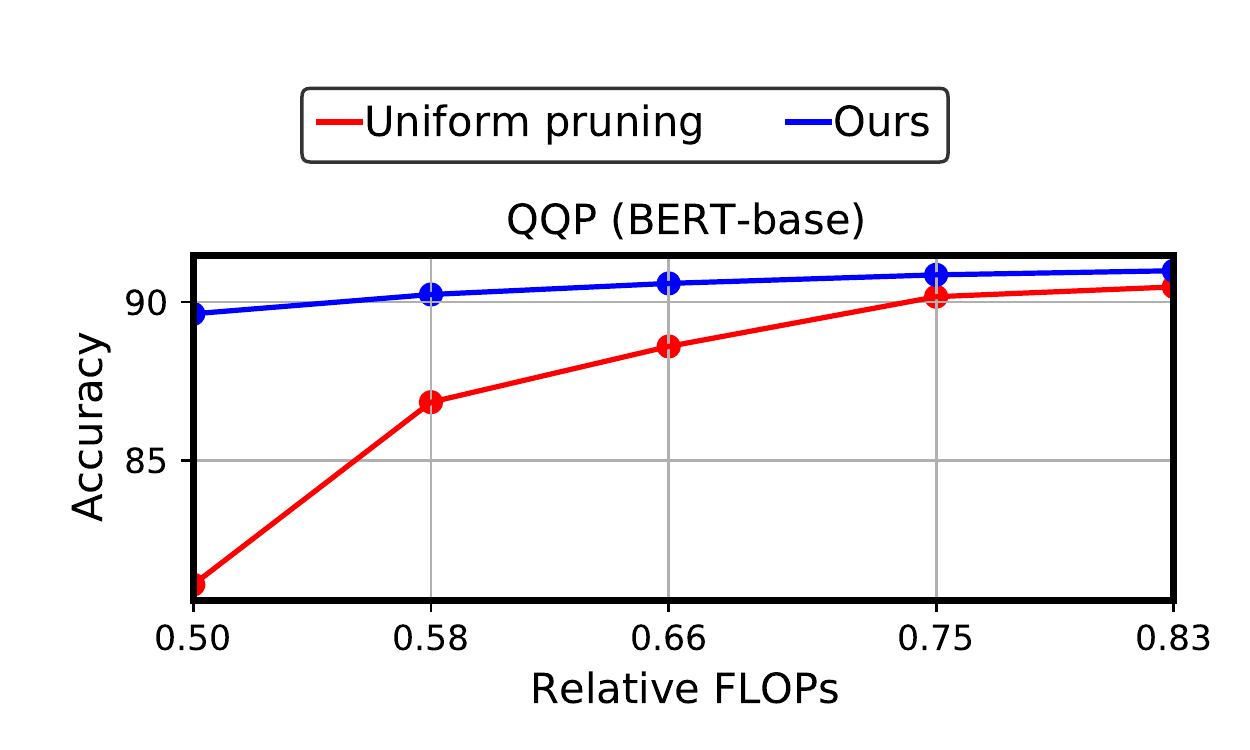}
\caption{
Performance of our method and Fisher-based uniform pruning. Mask tuning is applied to both methods.
}
\label{fig:ablation3}
\end{figure*}

To show the effectiveness of the mask search and re-arrangement stages, we compare the performance of our method with uniform Fisher pruning, which prunes every layer with the same sparsity.
Mask tuning is applied to both methods.
\fref{fig:ablation3} shows that the accuracy of our method significantly outperforms that of uniform pruning by up to 8.6\%. The result demonstrates the necessity of our mask search and re-arrangement techniques in finding quality binary masks.

\subsection{Societal Impacts} 

We believe our work would not bring immediate negative impacts on the society, as it aims to accelerate the model inference without affecting the output quality.
Our work can partly relieve the environmental concern due to DNN training, as it eliminates the need of re-training after pruning.

\end{document}